\documentclass[10pt,journal,compsoc]{IEEEtran}

\ifCLASSOPTIONcompsoc
  \usepackage[nocompress]{cite}
\else
  \usepackage{cite}
\fi

\usepackage{graphicx}
\usepackage[pagebackref,breaklinks,colorlinks]{hyperref}
\ifCLASSINFOpdf
\else
\fi

\usepackage{amsmath}
\usepackage{amssymb}
\usepackage{booktabs}
\usepackage{colortbl}
\usepackage[table]{xcolor}
\usepackage{color}

\usepackage{bm}
\usepackage{nicefrac} 

\usepackage{xspace}
\providecommand{\eg}{\textit{e.g.}\@\xspace}
\providecommand{\ie}{\textit{i.e.}\@\xspace}

\newlength\savewidth

\usepackage{enumitem}

\usepackage{algorithmic}

\usepackage{mathtools}

\usepackage{orcidlink}

\usepackage{xspace}
\makeatletter
\DeclareRobustCommand\onedot{\futurelet\@let@token\@onedot}
\def\@onedot{\ifx\@let@token.\else.\null\fi\xspace}

\def\eg{\emph{e.g}\onedot} 
\def\ie{\emph{i.e}\onedot}

\makeatother

\makeatletter
\renewcommand{\paragraph}{%
 \@startsection{paragraph}{4}%
 {\z@}{0.5em}{-1em}%
 {\normalfont\normalsize\bfseries}%
}
\makeatother

\usepackage[capitalize]{cleveref}
\crefname{section}{Sec.}{Secs.}
\Crefname{section}{Section}{Sections}
\Crefname{table}{Table}{Tables}
\crefname{table}{Tab.}{Tabs.}

\usepackage{amsmath,amsfonts,bm}

\def\eqref#1{equation~\ref{#1}}

\def\1{\bm{1}}

\DeclareMathAlphabet{\mathsfit}{\encodingdefault}{\sfdefault}{m}{sl}
\SetMathAlphabet{\mathsfit}{bold}{\encodingdefault}{\sfdefault}{bx}{n}

\newcommand{\nerf}{NeRF\@\xspace}
\newcommand{\nerfs}{NeRFs\@\xspace}
\newcommand{\siren}{{\sc siren}\xspace}

\newcommand{\pigan}{pi-GAN\xspace}

\newcommand{\egtd}{EG3D\xspace}
\newcommand{\egtdc}{EG3D~\cite{eg3d}\xspace}

\newcommand{\dfrc}{Deep3DFaceRecon~\cite{deng2019accurate}\xspace}
\newcommand{\ganctrl}{GAN-Control\xspace}
\newcommand{\ganctrlc}{GAN-Control~\cite{shoshan2021gancontrol}\xspace}
\newcommand{\disco}{DiscoFaceGAN\xspace}
\newcommand{\discoc}{DiscoFaceGAN~\cite{deng2020disentangled}\xspace}

\newcommand{\headnerfc}{HeadNeRF~\cite{hong2021headnerf}\xspace}

\newcommand{\pose}{\xi}
\newcommand{\coord}{\mathbf{x}}
\newcommand{\noise}{\mathbf{z}}
\newcommand{\tnoise}{\Tilde{\mathbf{z}}}
\newcommand{\dir}{\mathbf{d}}
\newcommand{\density}{\sigma}
\newcommand{\col}{\mathbf{c}}

\newcommand{\zshp}{\mathbf{z}_\text{shape}}
\newcommand{\zexp}{\mathbf{z}_\text{exp}}

\newcommand{\tzshp}{\Tilde{\mathbf{z}}_\text{shape}}
\newcommand{\tzexp}{\Tilde{\mathbf{z}}_\text{exp}}
\newcommand{\tztex}{\Tilde{\mathbf{z}}_\text{tex}}
\newcommand{\tzels}{\Tilde{\mathbf{z}}_\text{else}}

\newcommand{\tdmm}{3DMM\xspace}

\newcommand{\generator}{G}
\newcommand{\map}{\mathcal{M}}

\newcommand{\discriminator}{D}

\newcommand{\recon}{R}

\newcommand{\absrp}{\sigma}

\newcommand{\numsamples}{N}

\newcommand{\expo}[1]{\exp\left(#1\right)}
\newcommand{\deltatime}{\delta}

\begin{document}
\title{CGOF++: Controllable 3D Face Synthesis with Conditional Generative Occupancy Fields}

\author{Keqiang Sun\orcidlink{0000-0003-2900-1202},
        Shangzhe Wu\orcidlink{0000-0003-1011-5963},
        Ning Zhang\orcidlink{0000-0002-4569-0397},
        Zhaoyang Huang\orcidlink{0000-0001-7688-1471},
        Quan Wang\orcidlink{0000-0001-6943-4569},
        Hongsheng Li\orcidlink{0000-0002-2664-7975},
\IEEEcompsocitemizethanks{\IEEEcompsocthanksitem Keqiang Sun, Zhaoyang Huang are with the Department of Electronic Engineering, Chinese University of Hong Kong, Hong Kong SAR. E-mail: {kqsun, drinkingcode}@link.cuhk.edu.hk, 
\protect

\IEEEcompsocthanksitem Shangzhe Wu is with the Department of Engineering Science, University of Oxford, OX1 2JD Oxford, U.K. E-mail: szwu@robots. ox.ac.uk,
\protect
\IEEEcompsocthanksitem Ning Zhang and Quan Wang are with SenseTime Research, Beijing, China. E-mail: {zhangning2, wangquan}@sensetime.com,
\protect
\IEEEcompsocthanksitem Hongsheng Li is with CUHK MMlab, Centre for Perceptual and Interactive Intelligence, Shanghai AI Laboratory. E-mail: hsli@ee.cuhk.edu.hk, 

}%
\thanks{Manuscript received November 09, 2022.}
}

\markboth{Journal of \LaTeX\ Class Files,~Vol.~14, No.~9, November~2022}%
{Shell \MakeLowercase{\textit{et al.}}: Bare Demo of IEEEtran.cls for Computer Society Journals}

\IEEEtitleabstractindextext{%
\begin{abstract}
Capitalizing on the recent advances in image generation models, existing controllable face image synthesis methods are able to generate high-fidelity images with some levels of controllability, e.g., controlling the shapes, expressions, textures, and poses of the generated face images.
However, previous methods focus on controllable 2D image generative models, which are prone to producing inconsistent face images under large expression and pose changes.
In this paper, we propose a new NeRF-based conditional 3D face synthesis framework, which enables 3D controllability over the generated face images by imposing explicit 3D conditions from 3D face priors.
At its core is a conditional Generative Occupancy Field (cGOF++) that effectively enforces the shape of the generated face to conform to a given 3D Morphable Model (3DMM) mesh, built on top of \egtdc, a recent tri-plane-based generative model.
To achieve accurate control over fine-grained 3D face shapes of the synthesized images, we additionally incorporate a 3D landmark loss as well as a volume warping loss into our synthesis framework.
Experiments validate the effectiveness of the proposed method, which is able to generate high-fidelity face images and shows more precise 3D controllability than state-of-the-art 2D-based controllable face synthesis methods.

\end{abstract}

}

\maketitle

\IEEEdisplaynontitleabstractindextext

\IEEEpeerreviewmaketitle

\IEEEraisesectionheading{\section{Introduction}\label{sec:introduction}}

\IEEEPARstart{R}{ecent} success of Generative Adversarial Networks (GANs)~\cite{goodfellow2014generative} has led to tremendous progress in face image synthesis.
State-of-the-art methods, such as BigGAN~\cite{biggan} and StyleGAN~\cite{karras2019style, Karras2020stylegan2, Karras2021}, are capable of generating photo-realistic face images.
Apart from photo-realism, being able to control the appearance and geometry of the generated facial images is also key in many real-world applications, such as face animation, reenactment, and free-viewpoint rendering.

Early works on controllable face synthesis rely on external attribute annotations to learn an attribute-guided face image generation model. For instance, StarGANs\cite{ choi2018stargan, choi2020stargan} construct domains using the attribute labels and introduce a domain-aware reconstruction loss for the conditioning of the generative network. Many other methods\cite{di2017face, lu2018attribute, wang2018attribute, shen2020interfacegan} also make use attributes annotations for conditional facial image synthesis.
However, these attributes, such as ``big nose'', ``chubby'' and ``smiling'' in CelebA dataset~\cite{liu2015faceattributes}, can only provide an abstract semantic-level condition on the generation, 
and the generated faces often lack 3D geometric consistency.
Moreover, it is often much harder to obtain low-level geometric annotations beyond semantic labels for direct 3D supervision.

\begin{figure*}[t]
\begin{center}
\includegraphics[trim={0 0 20px 0}, clip, width=1\linewidth]{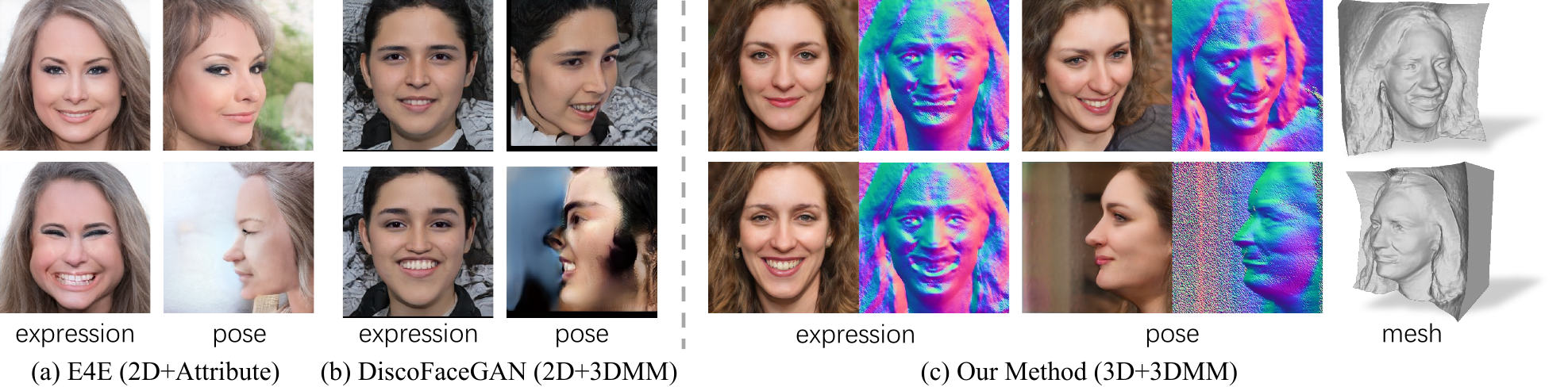}
\end{center}
\caption{
Existing controllable face synthesis methods, such as (a) E4E~\cite{e4e} and (b) DiscoFaceGAN~\cite{deng2020disentangled}, fail to preserve consistent geometry in the face images generated with large expression and pose variations, due to the lack of a 3D representation.
(c) Our proposed \emph{controllable 3D face synthesis} method leverages a \nerf-based 3D generative model conditioned on a prior 3DMM,
which enables precise and disentangled 3D control over the generated face images.
}
\label{fig:teaser}
\vspace{-3pt}
\end{figure*}

Recently, researchers attempted to incorporate 3D priors from parametric face models, such as 3D Morphable Models (3DMMs)~\cite{blanz1999morphable, bfm09}, into StyleGAN-based synthesis models, allowing more precise 3D control over the generated images. DiscoFaceGAN~\cite{deng2020disentangled} introduces the imitative loss between the rendered 3DMM image and the generated image. StyleRig~\cite{tewari2020stylerig} and PIE~\cite{tewari2020pie} devise an exquisite architecture to benefit from the 3DMM for latent code disentanglement. GAR\cite{piao2021inverting} proposes to embed 3DMM normal map into the StyleGAN~\cite{karras2019style} to control the geometry of the generated image.
However, despite their impressive image quality, these models still tend to produce \emph{3D inconsistent} faces under large expression and pose variations due to the lack of 3D representations, as shown in~\cref{fig:teaser}.

With the advances in differentiable rendering and neural 3D representations, a recent line of work has explored photo-realistic 3D face generation using only 2D image collections as training data~\cite{Szabo:2019, pigan, shi21lifting, gu2022stylenerf, orel2021styleSDF, eg3d, xu2021gof, deng2021gram, xu2021volumegan}, where they generate not only 2D images, but the implicit 3D geometry as well.
Neural Radiance Fields (NeRFs)~\cite{mildenhall2020nerf}, in particular, have enabled 3D generative models, such as pi-GAN~\cite{pigan}, EG3D~\cite{eg3d} and StyleNeRF~\cite{gu2022stylenerf}, to synthesize high-fidelity, 3D consistent faces, by training with 2D images only.
However, these models are purely generative and do not support precise 3D control over the generated faces, such as facial expressions.

In this work, we aim to connect these two groups of research---controllable face synthesis and 3D generative models---and present a NeRF-based 3D conditional face synthesis model that can generate high-fidelity face images with precise 3D control over the 3D shape, expression, and pose of the generated faces, by leveraging a parametric 3DMM face model.
Imposing precise 3D conditions on an implicit neural radiance field is non-trivial.
A naive baseline solution is to enforce that the input 3DMM parameters can be reproduced from the generated face image via a 3DMM reconstruction model, similar to~\cite{deng2020disentangled}.
Unfortunately, this 3DMM parameter reconstruction loss only provides \emph{indirect} supervisory signals to the underlying \nerf volume, and is insufficient for achieving precise 3D control, as shown in \cref{fig:ablation}.%

We seek to impose explicit 3D conditions \emph{directly} on the \nerf volume.
To this end, we propose conditional Generative Occupancy Field (cGOF++).
It consists of a mesh-guided volume sampling procedure and a complimentary depth-aware density regularizer. The mesh-guided volume sampler gradually concentrates the sampled position around the given mesh, whereas the density regularizer suppresses the density value of the far-away position according to their distance to the given 3DMM surface.
To achieve fine-grained 3D control, such as expression, we further introduce a 3D landmark loss and a volume warping loss. 
We develop the model on top of \egtdc, a high fidelity \nerf-based method.
We recurrent many existing methods and conduct the comparison with them in terms of the disentangle score. The proposed method outperforms existing methods by a large margin, which demonstrates the accurate controllability of the proposed method. Detailed ablation study further demonstrates the efficiency of the proposed components.
Moreover, we study to inverse cGOF++ to fit in-the-wild images and manage to edit the fitted images.

The main contributions are as follows.

1) To the best of our knowledge, we are one of the first works to propose the task of controllable 3D face generation. We also set up the metrics and build a comprehensive benchmark for the comparison in this task.

2) Technically, besides the baseline with a simple reconstruction model, we further propose the Mesh Guided Sampler and depth-aware density regularizer, which significantly reduce the Chamfer Distance by $61.28\%$ and promote the Landmark Correlation by $49.04\%$, indicating the accurate control over the 3D space.

3) Our proposed method outperforms \textbf{all} the controllable face synthesis methods, including the concurrent methods, in terms of \textbf{all} disentangle scores and landmark errors, demonstrating the superiority of our approach.

This article is an extension of our previous work~\cite{cgof}. In this article, we reimplement the proposed method on top of a more recent NeRF-based generative model \egtd~\cite{eg3d} and train with the {Flickr-Faces-HQ (FFHQ)}~\cite{karras2019style} to obtain the high-fidelity controllable generative model. This demonstrates the generalizability of the proposed method and significantly improves the visual quality, as well as the geometry fidelity. We update all the qualitative and quantitative results, provide more technical details, and include additional experiments and the expanded literature review.

\section{Related Work}

In order to assess our contribution in relation to the vast literature on controllable image generation, it is important to consider three aspects of each approach: how much 3D information is used in the generative network, which factor is controllable, and where does the controllability comes from. Below and in \cref{tab:related_work} we compare our contribution to prior works based on these factors.

\subsection{3D-Aware GAN}
Another line of work in 3D-Aware generative models has looked into disentangling 3D geometric information from 2D images, and learn to generate various 3D structures, such as voxels~\cite{wu2016learning, zhu2018visual, Gadelha:2017, NguyenPhuoc2019, nguyen2020blockgan, lunz2020inverse}, meshes~\cite{Szabo:2019, Liao2020CVPR}, and Neural Radiance Fields~(\nerfs)~\cite{pigan, eg3d, gu2022stylenerf, schwarz2020graf, xu2021volumegan, niemeyer2021giraffe, niemeyer2021campari}, by training on image collections using differentiable rendering. 
The key idea is to use a discriminator that encourages images rendered from random viewpoints sampled from a prior pose distribution to be indistinguishable from real images.
Among these works, SofGAN~\cite{chen2021sofgan} ensures viewpoint consistency via the proposed semantic occupancy field but requires 3D meshes with semantic annotations for pre-training.
\pigan~\cite{pigan} has demonstrated the unsupervised high-fidelity 3D synthesis results adopting a \siren-based \nerf representation as the generator.
Many recent works~\cite{gu2022stylenerf, deng2021gram, orel2021styleSDF, xu2021volumegan, eg3d} further enhanced the image quality significantly. StyleNeRF~\cite{gu2022stylenerf} and VolumeGAN~\cite{xu2021volumegan} introduce the upsampler to translate the blurry rendered image to a super-resolution image, while maintaining the geometrical consistency. Instead of upsampling the generated image with a neural network, GRAM~\cite{deng2021gram} and StyleSDF~\cite{orel2021styleSDF} propose more efficient representations to enhance the geometry fidelity.
\egtdc introduces a novel efficient representation, \ie tri-plane, and makes use of the neural renderer for super-resolution, where a dual-discrimination keeps the consistency between the rendered raw image and the super-resolution image. Hence \egtdc currently has the best visual quality, as well as the most vivid geometry.
In this work, we build on top of \egtdc and directly learn from unlabeled image collections with a prior \tdmm.

\subsection{Controllable Generative Adversarial Networks}
Generative Adversarial Nets (GANs)~\cite{goodfellow2014generative} gained popularity over the last decade due to their remarkable image generation ability~\cite{radford2016unsupervised, kanazawa2018learning, biggan, karras2019style, Karras2020stylegan2}. Prior works have studied disentangled representation learning in generative models.
Early works tend to make use of labeled data with find-grained attribute annotations.
For example, a line of works~\cite{di2017face, lu2018attribute, wang2018attribute, choi2018stargan, choi2020stargan, shoshan2021gancontrol} proposes to train a conditional generative model, which takes in an attribute label and yields corresponding images, by constructing attribute domains first and then supervising the network training with the labeled images. 
Instead of training a conditional generative model, the other line of works~\cite{harkonen2020ganspace, shen2020interfacegan, zhu2020domain, zhuang2021enjoy, shen2021closed, li2021image} manages to explore the latent space of pretrained generative models, which can randomly generate high-fidelity images of a specific category, to find directions of certain attributes, along which they are capable of controlling the attributes of the generated images.
Many 2D-based representations like
2D key points~\cite{wayne2018reenactgan, zakharov2019few, huang2020learning, ha2020marionette} and 
semantic maps~\cite{lee2020maskgan, ling2021editgan, zhu2020sean, kim2021exploiting, zhu2021barbershop} are used to control the shape of the synthesised image.
Many recent works~\cite{tewari2020stylerig, tewari2020pie, deng2020disentangled, FreeStyleGAN2021, piao2021inverting} propose to introduce 3D prior to the controllable facial image synthesis to enhance the control effectiveness.
However, these existing works on controllable facial image synthesis, including the fore-mentioned ones still rely on 2D-image-based generative models, which do not guarantee the 3D consistency of the generated images.

\subsection{Controllable 3D Generative Models}
Recently, various methods have been proposed to generate 3D portraits and bodies.

To learn controllable 3D face synthesis models, researchers have leveraged prior face models.
3D Morphable Models~(3DMMs)~\cite{blanz1999morphable, bfm09} is a widely-used parametric face model, which represents human face as a linear combination of a set of principle components derived from 3D scans via Principle Components Analysis (PCA).
Many previous works~\cite{deng2020disentangled, tewari2020stylerig, chen2021sofgan, piao2021inverting} have proposed to incorporate 3D priors of 3DMM into generative models to enable controllable 3D face synthesis.
In particular, DiscoFaceGAN~\cite{deng2020disentangled} proposes the imitative loss to encourage the generated image to be consistent with the rendered image.
StyleRig~\cite{tewari2020stylerig} and PIE~\cite{tewari2020pie} devise an exquisite training process to obtain a disentangled StyleGAN~\cite{karras2019style} latent code editor.
GAR~\cite{piao2021inverting} introduces a Renderer Block where 3DMM normal maps are embedded into the image feature maps to synthesis images, which allow the generated images containing more 3D information.
Although these methods incorporate 3D priors, they still reply on 2D-based generator for the final image generation. HeadNeRF~\cite{hong2021headnerf} brings a 3D \nerf-based representation and 3DMM together, but it is a reconstruction model rather than a generative model, trained with an image reconstruction loss using annotated multi-view datasets.

Some more recent works~\cite{tang20223dfaceshop, yue2022anifacegan, bergman2022gnarf, zhang2023avatargen} propose to make a 3D-based generator controllable. 3DFaceShop~\cite{tang20223dfaceshop} achieves control over facial shape and expression by applying semantic mask supervision on the generated image, rather than directly supervising on the 3D volume space. As a result, this approach is inefficient for 3D face control.
AnifaceGAN~\cite{yue2022anifacegan} mimics mesh deformation to achieve direct control of the 3D shape, but also suffers from inaccuracies due to the lack of a clear and condensed surface for warping the implicit volume according to the deformation of two meshes.
For body generation, GNARF~\cite{bergman2022gnarf} and AvatarGen~\cite{zhang2023avatargen} generate human bodies or faces in the canonical space and deform them to the target gesture, but they also rely on mesh deformation to warp the whole volume, which is inaccurate due to the lack of a condensed surface.

In contrast, our proposed method, cGOF++, encourages the volume to compress to the surface near the input mesh, providing a solid foundation for following auxiliary fine-grained controlling losses. This allows us to control the 3D shape effectively and generate high-quality facial images.

\begin{table}[t]
\caption{\label{tab:related_work}Comparison with selected prior work: Dimension, Controllable Factors, and Condition Type}
\centering
\begin{tabular}{cccc}
\toprule
Method                      & Dims     & Ctrl. Fac.  & Cond. Type \\
\midrule
\cite{karras2019style, Karras2020stylegan2}
                            & 2D            & -             & -  \\
\midrule
\cite{di2017face, lu2018attribute, wang2018attribute, choi2018stargan, choi2020stargan, shoshan2021gancontrol}
                            & 2D            & A.          & A.A.\\
\cite{harkonen2020ganspace, shen2020interfacegan, zhu2020domain, zhuang2021enjoy, shen2021closed, li2021image}
                            & 2D            & L.       & A.A.\\
\cite{wayne2018reenactgan, zakharov2019few, huang2020learning, ha2020marionette} & 2D & P. & P.A.\\
\cite{lee2020maskgan, ling2021editgan, zhu2020sean, kim2021exploiting, zhu2021barbershop}       & 2D            & S.       & S.A.  \\
\cite{deng2020disentangled, tewari2020stylerig, tewari2020pie, piao2021inverting}
                            & 2.5D          & M.F.    & M.\\
\cite{chen2021sofgan, sun2022fenerf, shi2022semanticstylegan, chen2022sem2nerf}       & 2.5D          & L.   & 3DS. \\
\midrule
\cite{pigan, eg3d, gu2022stylenerf, orel2021styleSDF, deng2021gram, xu2021volumegan}
                            & 3D            & -             & -     \\
Ours                        & 3D            & M.F.    & M.      \\
\bottomrule
\end{tabular}
{A.: Attributes, A.A.: Attribute Annotations, L.: Latent code, P.: Key Points, P.A.: Key Point Annotations, S.: Semantic Map, S.A.: Semantic Annotations, M.: 3D Morphable Model (3DMM), M.F.: 3DMM Factors, 3DS.: 3D Semantic Map, -: Not Controllable.
}
\end{table}

\section{\label{sec_method}Method}

\begin{figure*}[t]
\begin{center}
\resizebox{1.0\linewidth}{!}{
\includegraphics[width=0.95\linewidth]{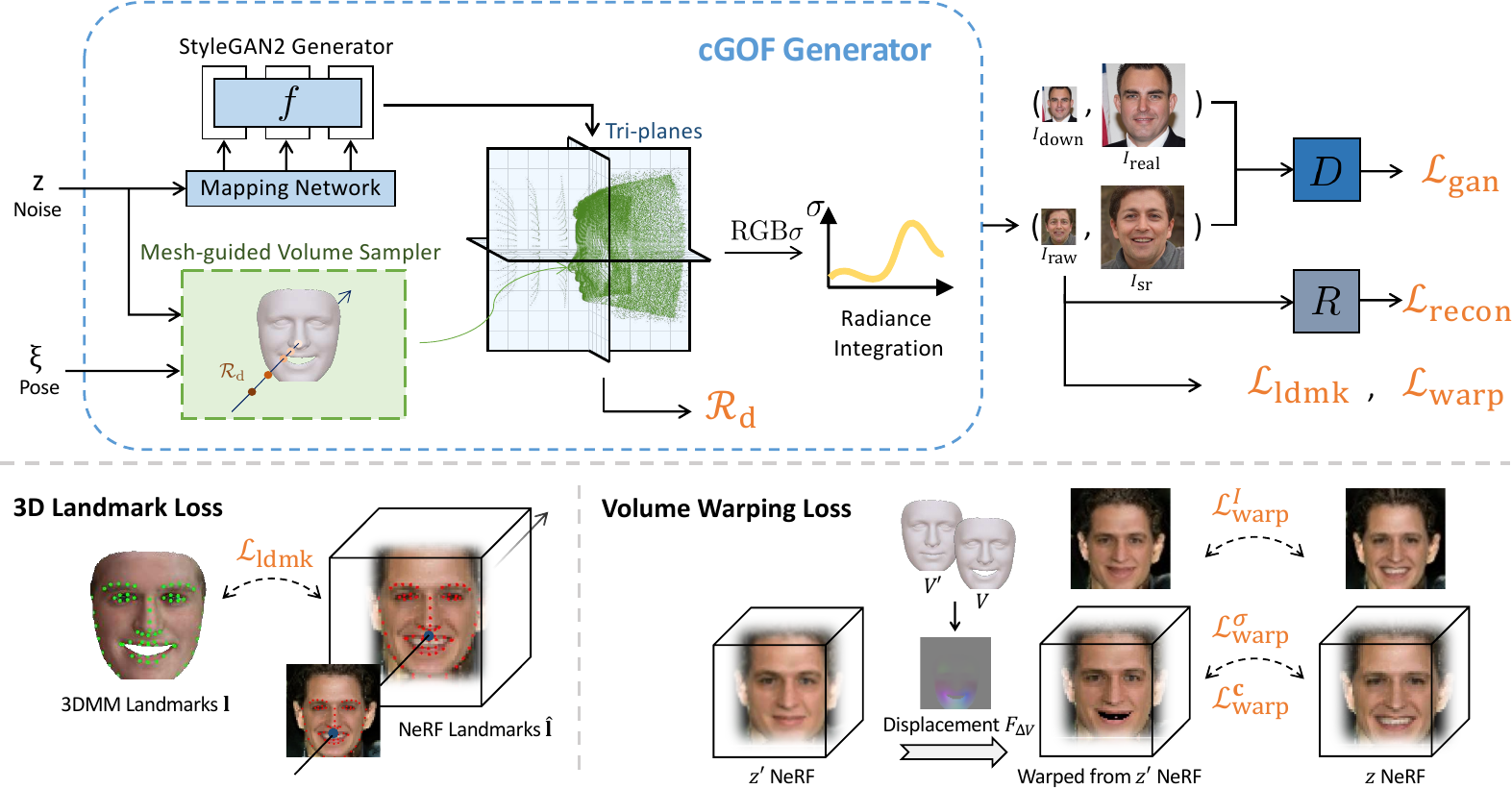}
}
\end{center}
\caption{Method overview. \textit{(Top)} Conditional Generative Occupancy Field (cGOF++) leverages a mesh-guided volume sampler which effectively conditions the generated \nerf on an input 3DMM mesh.
It is trained in an adversarial learning framework using only single-view images.
\textit{(Bottom)} The 3D landmark loss encourages the semantically important facial landmarks to follow the input mesh, and the volume warping loss enforces two NeRF volumes generated with different expression codes to be consistent through a warping field induced from the corresponding 3DMM meshes.
}
\label{fig:framework}
\vspace{-10pt}
\end{figure*}

Given a collection of single-view real-world face images $\mathcal{Y}$, where each face is assumed to appear only once without multiple views, the goal of this work is to learn a controllable 3D face synthesis model that can generate face images with desired 3D shape, expression, and pose.
To achieve this goal, we propose a \nerf-based conditional 3D generative model $G$ and incorporate priors from a parametric 3D Morphable Model (\tdmm)~\cite{blanz1999morphable}.
At the core of our method is a novel conditional Generative Occupancy Field (cGOF++) as well as two auxiliary volumetric losses that enable effective 3D control based on the \tdmm, as illustrated in Figure~\ref{fig:framework}.

\subsection{Preliminaries}

\subsubsection{Neural Radiance Field (NeRF)}
\nerf~\cite{mildenhall2020nerf} represents a 3D scene as a radiance field parametrized by a Multi-Layer Perceptron (MLP) that predicts a density value $\density \in \mathbb{R}$ and a radiance color $\col \in \mathbb{R}^3$ for every 3D location, given its $xyz$ coordinates and the viewing direction as input.
To render a 2D image of the scene from an arbitrary camera viewpoint $\pose$, we cast rays through the pixels into the 3D volume, and evaluate the density values and radiance colors at a number of sampled points along each ray.
Specifically, for each pixel, the camera ray $\mathbf{r}$ can be expressed as $\mathbf{r}(t) = \mathbf{o} + t \dir$, where $\mathbf{o}$ denotes the camera origin, $\dir$ the direction, and $t$ defines a sample point within near and far bounds $t_\text{n}$ and $t_\text{f}$.
For each 3D sample point $\mathbf{x}_i$, we query the \nerf network $f$ to obtain its density value $\density_i$ and radiance color $\col_i$ as $(\density_i, \col_i) = f(\mathbf{x}_i, \dir)$.
Using volume rendering~\cite{max1995optical, mildenhall2020nerf}, the final color of the pixel is given by
\vspace{-5pt}
\begin{align}
\label{eqn:render_coarse}
\begin{split}
    \hat{C}(\mathbf{r}) & = \sum_{i=1}^{\numsamples}T_i (1-\expo{-\absrp_i \deltatime_i}) \mathbf{c}_i\,,
    \quad\\
    &\textrm{ where } \quad
    T_i = \exp(- \sum_{j=1}^{i-1} \absrp_j \deltatime_j)\,,
\end{split}
\end{align}
and $\deltatime_i = t_{i+1} - t_i$ is the distance between adjacent samples.
To optimize a \nerf of a 3D scene, the photometric loss is computed between the rendered pixels and the ground-truth images captured from a dense set of viewpoints.

\subsubsection{Occupancy Field (OF)}
The OF~\cite{xu2021gof} is similar to the NeRF~\cite{mildenhall2020nerf}, except that the occupancy value is binary (either $0$ or $1$) while the density value in NeRF is contiguous.

On the one hand, OF is similar to the NeRF.
As discussed in the GOF~\cite{xu2021gof}, the alpha value ($\alpha$), which resembles the occupancy value in OF, is determined by the density ($\sigma$), indicating these two representations are similar and can be interconverted.
On the other hand, NeRF have relatively continuous density values, while the occupancy value is typically binary, indicating the occupied region has extremely large densities and the empty region has small densities.

\subsubsection{Generative Radiance Field}
Several follow-up works have extended this representation to generative models~\cite{pigan, eg3d, gu2022stylenerf, deng2021gram, xu2021gof}.
Instead of using a multi-view reconstruction loss, they train a discriminator that encourages images rendered from randomly sampled viewpoints to be indistinguishable from real images, allowing the model to be trained with only single-view image collections.

Our method is built on top of \egtd~\cite{eg3d},
a generative radiance field model based on a tri-plane-based \nerf representation.
During training, images are rendered from random viewpoints $\pose$ sampled from the training dataset.
To model geometry and appearance variations, a random noise code $\noise$ is sampled from a standard normal distribution and mapped to the $W$-space as a number of intermediate latent scalars through a mapping network.
These intermediate latent scalars are used to modulate the StyleGAN2 generator, which synthesizes the tri-plane features.
After sampling points in the space according to the viewpoints $\pose$, the corresponding features are fetched from the tri-planes and processed with a lightweight decoder to predict the color and density value. The volume rendering module then integrates the point color and density along each camera ray to render the low-resolution image, which is then translated to a corresponding high-resolution image with a super-resolution module.

The rendered images $I_\text{g} = \generator(\noise, \pose)$ and the real images $I$ sampled from the training dataset with the same viewpoint $\pose$ are then passed into the StyleGAN2 Discriminator $\discriminator$, and the model is trained using a non-saturating GAN loss with R1 regularization following~\cite{gan_convergence}:
\begin{align}
\label{eqn:eqlabel}
\begin{split}
\mathcal{L}(\theta) &=
\mathbb{E}_{\noise\sim p_z, \xi\sim p_\xi}[f(\discriminator(\generator(\noise, \pose)))]\\
&+ \mathbb{E}_{I\sim p_\mathcal{D}}[f(-\discriminator(I)) + \lambda \lvert \nabla \discriminator (I) \rvert ^ 2],
\end{split}
\end{align}
where $f(u) = -\log(1 + \exp(-u))$ and $\lambda$ is a hyperparameter balancing the regularization term.

\subsection{Controllable 3D Face Synthesis}

In order to learn a controllable 3D face synthesis model, we leverage priors of a \tdmm~\cite{bfm09}, which is a parametric morphable face model, with the shape, expression, pose and other factors of a face being modeled by a set of Principal Component Analysis (PCA) bases and coefficients $\tnoise = (\tzshp, \tzexp, \tztex, \tzels)$.
Our goal is to condition the generated face image $I_\text{g}$ on a set of 3DMM parameters $\tnoise$ as well as a camera pose $\xi$, which specify the desired configuration of the generated face image.

\subsubsection{Baseline.}
To enforce such conditioning, following \cite{deng2020disentangled}, we make use of an off-the-shelf pretrained and fixed 3DMM reconstruction model $\recon$ that predicts a set of 3DMM parameters from a single 2D image~\cite{deng2019accurate}.
This allows us to incorporate a 3DMM parameter reconstruction loss that ensures the generated face images to commit to the input condition.

Specifically, we sample a random noise $\noise \in \mathbb{R}^d$ from a standard normal distribution $p_\noise$, which is later fed into the \egtd generator $\generator$ to produce a face image $I_\text{g}$.
Unlike \cite{deng2020disentangled}, which additionally trains a set of VAEs to map the noise $\noise$ to meaningful 3DMM coefficients $\tnoise$, we assume that the 3DMM coefficients of all the training images follow a normal distribution and simply standardize them to obtain the corresponding normalized noise code 
$\noise = \tau(\tnoise) = \tilde{L}^{-1}(\tnoise - \tilde{\mu})$, where $\tilde{\mu}$ and $\tilde{L}$ are the mean and Cholesky decomposition component of covariance matrix of the 3DMM coefficients estimated from all the training images using the 3DMM reconstruction model $\recon$.

During training, we predict 3DMM parameters from the generated image using the reconstruction model $\recon$, normalize it and minimize a reconstruction loss between the predicted normalized parameters and the input noise:
\begin{align}
\label{eqn:l_recon}
\mathcal{L}_\text{recon} = \| \hat{\noise} - \noise \|_1,
\quad \text{where} \quad \hat{\noise} = \tau(\recon(\generator(\noise, \pose))).
\end{align}

This results in a baseline generative model, where the conditioning is implicitly imposed through the 3DMM parameter reconstruction loss. Example results of this baseline model can be seen in \cref{fig:ablation}.

\subsubsection{\label{challenge}Challenges}
There are several issues with this baseline model.
First, the effectiveness of the conditioning depends heavily on the accuracy and generalization performance of the 3DMM reconstruction model $\recon$.
Fundamentally, these parameters only describe an extremely abstract, semantic representation of human faces, which is insufficient to govern the fine-grained geometric details in real faces precisely.
Since our model generates a 3D representation for rendering, a more effective solution is to impose the conditioning directly on the 3D volume, instead of just on the rendered 2D images.

\subsection{Conditional Generative Occupancy Field}
In order to impose precise 3D control on the \nerf volume, we first identify an effective way of enforcing the volume density to converge to the surface of human faces.
Inspired by \cite{xu2021gof}, we propose the conditional Generative Occupancy Field~(cGOF++), which consists of a mesh-guided volume sampler~(MgS) and a distance-aware volume density regularizer to compress the target space to the desired surface.

The key idea is to use the conditioning 3DMM mesh to explicitly guide the volume sampling procedure.
Given a random noise $\noise$, we first obtain the corresponding 3DMM coefficients $\tnoise = \tau^{-1} (\noise)$ by denormalization, which consists of $(\tzshp, \tzexp, \tztex, \tzels)$.
With these coefficients, we can sample a 3D face mesh $M_\text{in}$ from the 3DMM $\Lambda$:
\begin{align}
\label{eqn:3dmm_mesh}
\begin{split}
    M_\text{in} &= \Lambda (\zshp, \zexp)\\
    &= M_\text{mean} + B_\text{shape} \cdot \tzshp + B_\text{exp} \cdot \tzexp,
\end{split}
\end{align}
where $M_\text{mean}$ is the mean shape, and $B_\text{shape}$ and $B_\text{exp}$ are the deformation bases for shape and expression in 3DMM respectively.
During rendering, this mesh will serve as 3D condition, allowing us to apply importance sampling around the surface region, and suppress the volume densities far away from the surface.

Specifically, for each pixel within the face region, we find its distance $t_\text{m}$ to the input 3DMM mesh $M_\text{in}$ along the ray direction by rendering a depth map, taking into account the perspective projection.
This allows us to sample points along the ray $\mathbf{r}(t) = \mathbf{o} + t \dir$ within a set of uniformly spaced intervals around the surface intersection point of the ray and the input mesh:
\begin{align}
\label{eqn:neighboringsampler}
\begin{split}
t_i \sim& \mathcal{U} \left[ t_\text{m} + (\frac{i-1}{N} - \frac{1}{2}) \delta, t_\text{m} + (\frac{i}{N} - \frac{1}{2}) \delta \right],
\;\\&\text{where} \; i = 1, 2, \dots, N_\text{surf},
\end{split}
\end{align}
where $N_\text{surf}$ is the number of sampled points, and $\delta$ is a parameter controlling the ``thickness'' of the sampling region.
We gradually shrink this margin from  $0.5$ to $0.05$ of the volume range during training.

Furthermore, we suppress the volume density in the space far away from the given mesh using a distance-aware volume density regularizer.
Specifically, we additionally sample a sparse set of points along each ray, and denote the absolute distances from these sampling points to the mesh surface as $d_i = | t_i - t_\text{m} |, i = 1, \dots, N_\text{vol}$.
The distance-aware volume density regularizer $R_\text{d}$ is defined as:
\vspace{-3pt}
\begin{align}
\label{eqn:density_suppression}
R_\text{d} = \sum_{i=0}^{N_\text{vol}} \sigma_i \cdot
\left [\exp\left (\alpha \cdot \max(d_i-\delta/2, 0)\right ) - 1 \right ],
\end{align}
\noindent{where} $\density_i$ is the volume density evaluated at those sample points, and $\alpha$ is an inverse temperature parameter (set to $20$), controlling the strength of the penalty response to increased distance $\Delta d_i$.
Note that this penalty only applies to sample points beyond the surface sampling range $\frac{\delta}{2}$.
The mesh-guided volume sampler together with the volume density regularizer effectively constrains the generated \nerf volume to follow closely the input 3DMM mesh, as shown in \cref{tab:ablation}.

\subsection{Auxiliary Losses for Fine-grained 3D Control}
So far, the model is able to generate 3D faces following the overall shape of the conditioning 3DMM mesh.
However, since 3DMM is only a low-dimensional PCA model of human faces without capturing the fine-grained shape and texture details of real faces, enforcing the generated 3D faces to fully commit to the coarse 3DMM meshes actually leads to over-smoothed shapes, and does not guarantee precise control over the geometric details, especially under large facial expression changes.
To enable precise geometric control, we further introduce two auxiliary loss terms.

\subsubsection{3D Landmark Loss} encourages the semantically important facial landmarks to closely follow the input condition.
Specifically, we identify a set of 3DMM vertices that correspond to $N_\text{k} = 68$ facial landmarks~\cite{sagonas2013300w,deng2019accurate}.
Let $\{\mathbf{l}_k\}_{k=1}^{N_\text{k}}$ denote the 3D locations of these landmarks on the input 3DMM mesh.
In order to find the same set of landmarks on the generated image, we extract 3D landmarks $\{\hat{\mathbf{l}}_k\}_{k=1}^{N_\text{k}}$ from the \emph{estimated} 3DMM mesh and project them onto the 2D image to obtain the 2D projections.
We then render the depth values of \nerf at these 2D locations using volume rendering and back-project them to 3D to obtain the 3D landmark locations of the generated volume ${\{\hat{\mathbf{l}}'_k\}}$. To avoid the impact of occlusion, we skip the facial contour landmarks (numbered from 1 to 17) when calculating loss.
The landmark loss is defined among these three sets of 3D landmarks:
\begin{align}
\label{eqn:l_ldmk}
\mathcal{L}_\text{ldmk} = \sum_{k=1}^{N_\text{k}} \| \hat{\mathbf{l}}_k - \mathbf{l}_k \|_1 + \sum_{k=18}^{N_\text{k}} \| \hat{\mathbf{l}}'_k - \mathbf{l}_k \|_1.
\end{align}

Adopting this 3D landmark loss further encourages the reconstructed 3D faces to align with the input mesh at these semantically important landmark locations, which is particularly helpful in achieving precise expression control, as validated in \cref{tab:ablation}.

\subsubsection{Volume Warping Loss} enforces that the \nerf volume generated with a different expression code should be consistent with the original volume warped by a warping field induced from the two 3DMM meshes.
More concretely, let $M = \Lambda (\zshp, \zexp)$ be a 3DMM mesh randomly sampled during training, $M' = \Lambda (\zshp, \zexp')$ be a mesh of the same face but with a different expression code $\zexp'$, and $V$ and $V'$ be their vertices respectively.
We first compute the vertex displacements denoted by $\Delta V = V' - V$, and render them into a 2D displacement map $F_{\Delta V} \in \mathbb{R}^{H \times W \times 3}$, where each pixel $F_{\Delta V}^{(u,v)}$ defines the 3D displacement of its intersection point on the mesh $M$.

For each pixel $(u,v)$ in the image $I_\text{g}$ generated from the first \nerf, we obtain the densities and radiance colors $(\density_i, \col_i) = f(\coord_i, \dir, \noise)$ of the $N_\text{surf}$ mesh-guided sample points $\coord_i$ described above.
The superscript $(u,v)$ is dropped for simplicity.
We warp all the 3D sample points along each ray using the same 3D displacement value rendered at the pixel $F_{\Delta V}$ to obtain the warped 3D points $\coord_i' = \coord_i + F_{\Delta V}$ (assuming that all the points are already close to the mesh surface).
We then query the second \nerf generated with $\zexp'$ at these warped 3D locations $\coord_i'$ to obtain another set of densities and radiance colors $(\density_i', \col_i') = f(\coord_i', \dir', \noise')$, and encourage them to stay close to the original ones.

In addition, we also enforce the images rendered from the two radiance fields to be consistent.
We integrate the radiance colors $\col_i'$ of the \emph{warped} points with their densities $\density_i'$ using \cref{eqn:render_coarse}, and obtain an image $\hat{I}_\text{g}$ (with all pixels).
This image $\hat{I}_\text{g}$ rendered from the second radiance field is encouraged to be identical to the original image $I_\text{g}$.
The final warping loss is thus composed of three terms:
\begin{align}
\label{eqn:l_warp}
\begin{split}
\mathcal{L}_\text{warp} &=
\beta_\text{d} \cdot \sum_i^{N_\text{surf}} \| \density_i' - \density_i \|_1\\
&+ \beta_\text{c} \cdot \sum_i^{N_\text{surf}} \| \col_i' - \col_i\|_1
+ \beta_\text{I} \cdot \| \hat{I}_\text{g} - I_\text{g}\|_1,
\end{split}
\end{align}
where $\beta_\text{d}$, $\beta_\text{c}$ and $\beta_\text{I}$ are the balancing weights.

The final training objective is a weighted summation of the mentioned losses, given by
\begin{align}
\label{eqn:l_all}
\begin{split}
\mathcal{L} &= \lambda_\text{gan} \mathcal{L}_\text{gan}
            + \lambda_\text{recon} \mathcal{L}_\text{recon}
            + \lambda_\text{d} R_\text{d}\\
            &+ \lambda_\text{ldmk} \mathcal{L}_\text{ldmk}
            + \lambda_\text{warp} \mathcal{L}_\text{warp},
\end{split}
\end{align}
where the $\lambda$'s are the weights for each term.

\begin{figure}[t!]
    \centering
    \includegraphics[width=\linewidth]{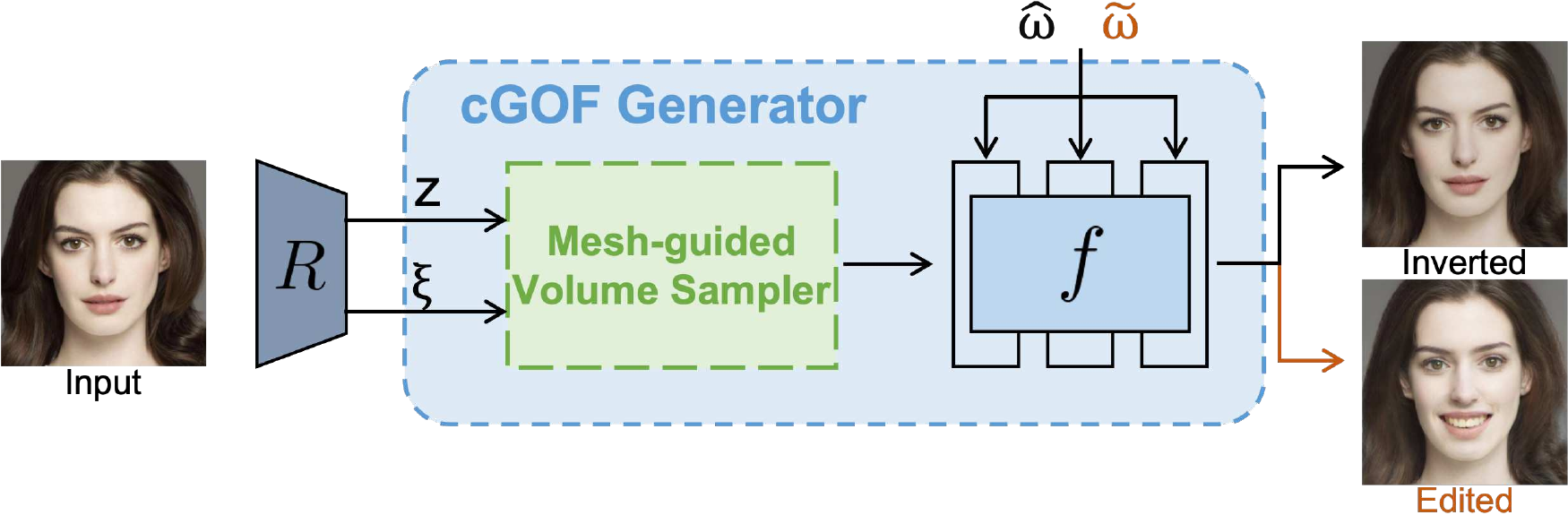}
    \caption{Process of image inversion and editing. Given an input image, we first use the face reconstruction network R to estimate the 3DMM coefficients $z$ and the camera pose $\pose$, and then apply PTI~\cite{pti} to obtain the predicted latent code $\hat{\omega}$. With an edited coda $\tilde{\omega}$, we manage to control the expression of the input image.}
    \label{Fig:inversion}
    \vspace{-5pt}
\end{figure}

\begin{table}[t]
\caption{Training details and hyper-parameter settings}\label{supmat:tab:param_set}
\vspace{-3pt}
\begin{center}
\begin{tabular}{lc}
\toprule
Parameter & Value/Range \\ \midrule
Optimizer & Adam \\
Generator learning rate & $2\times 10^{-3}$ \\
Discriminator learning rate & $2\times 10^{-3}$ \\
œNumber of iterations & $40,000$ \\
Batch size & 96\\
$N_\text{surf}$ & 48 \\
$N_\text{vol}$ & 48 \\
training image resolution & $512\times 512$ \\

\midrule
Loss weight $\lambda_{\text{gan}}$ & $1$ \\
Loss weight $\lambda_{\text{recon}}$ & $4$ \\
Loss weight $\lambda_{\text{d}}$ & $100$ \\
Loss weight $\lambda_{\text{ldmk}}$ & $20$ \\
Loss weight $\lambda_{\text{warp}}$ & $20$ \\

\midrule
$\noise$ & $\mathcal{N}^{411}(0, 1)$\\
Ray length & $(2.25, 3.30)$ \\
Average Camera Radius & $2.7$ \\
Field of view (FOV) & $13.373^\circ$\\
\bottomrule
\end{tabular}
\end{center}

\end{table}

\subsection{\label{inv_edit}Inversion and Editing}
To equip the proposed model with the capability to edit input facial images, we devise the inversion process for our proposed cGOF++, as shown in~\cref{Fig:inversion}. Given an in-the-wild image, we first use the face reconstruction network $\recon$ to estimate the \tdmm coefficients $\noise$ and the camera pose $\pose$, which determines the sampled position in the Mesh-guided Volume Sampler.

Experimentally, we find it hard to directly obtain the inverted \tdmm coefficients, while it is easier to invert a given image to the $W$-space code, the output of the mapping network in the framework~\cite{eg3d, karras2019style}. We first estimate the \tdmm coefficients $\noise$ of the given image with a pretrained face reconstruction network~\cite{deng2019accurate}.
Then we apply the pivotal tuning inversion (PTI)~\cite{pti} to optimize the inversed $W$-space latent code $\hat{\omega}$, as well as the generator parameters, to reconstruct the input image. 

In the editing process, based on the inversion results, we modify the reconstructed coefficient code $\noise$ to $\noise'$, and map them to the $W$-space as $\omega$ and $\omega'$. Inspired by DiscoFaceGAN~\cite{deng2020disentangled}, we edit the predicted latent code $\hat{\omega}$ with the difference of $\omega$ and $\omega'$.
The final edited latent code $\tilde{\omega}$ is thus
\begin{align}
\label{eqn:edit_w}
\begin{split}
\tilde{\omega} &= \hat{\omega} - \omega + \omega'
,
\end{split}
\end{align}
\noindent
where
$\omega = \map(\noise)$, $\omega' = \map(\noise')$, and $\map$ denotes the mapping network of the generator. The edited image $I_\text{edit}$ can thus be synthesised as
\begin{align}
\label{eqn:gen_edit}
\begin{split}
I_\text{edit} = \generator(z', \tilde{\omega}, \pose).
\end{split}
\end{align}

\subsection{\label{sec_expimpl}Implementation Details}
We build our model on top of the implementation of \egtd~\cite{eg3d}, which uses a tri-plane-based \nerf representation. We use the same architectures for the generator and discriminator, as well as the main training process in from the \egtd~\cite{eg3d}. 
We sample $N_\text{vol}=48$ coarse points evenly in the volume to obtain $N_\text{fine}=48$ fine points as introduced in~\cite{eg3d}.
Then we replace the fine points with $N_\text{surf} = 48$ surface points sampled around the 3DMM input mesh for rendering and optimize the $\generator$ with gradient back-propagation.

To implement our proposed conditional Generative Occupancy Field (cGOF++), we integrate a deep 3D face reconstruction model \dfrc, a weakly supervised face reconstruction model, to reconstruct 3DMM parameters from the generated images for the 3DMM reconstruction parameter loss $\mathcal{L}_\text{recon}$.
We follow~\cite{deng2019accurate} to adopt the popular 2009 Basel Face Model~\cite{bfm09} for shape and texture bases, and use the expression bases of~\cite{guo2018cnn}, built from FaceWarehouse~\cite{cao2013facewarehouse}. The expression components are the PCA model of the offsets between the expression meshes and the neutral meshes of individual persons.

The final model is trained for 168 hours on 8 NVIDIA V100 Tensor Core GPUs.
To determine loss weights $\lambda$s, we start from the original \egtd~\cite{eg3d} and add the proposed components one by one, as in \cref{tab:ablation}. For each component, we first initialize the hyperparameter lambda as $1$, and empirically find a reasonable value before taking a fine-grained search. We summarize the detailed hyper-parameters in \cref{supmat:tab:param_set}.

\section{Experiments}

\begin{figure*}
\center
\setlength\tabcolsep{0pt}
{
\renewcommand{\arraystretch}{0.0}
\footnotesize
\setlength{\tabcolsep}{3pt}{
\begin{tabular}{cc}
    \includegraphics[width=0.47\linewidth]{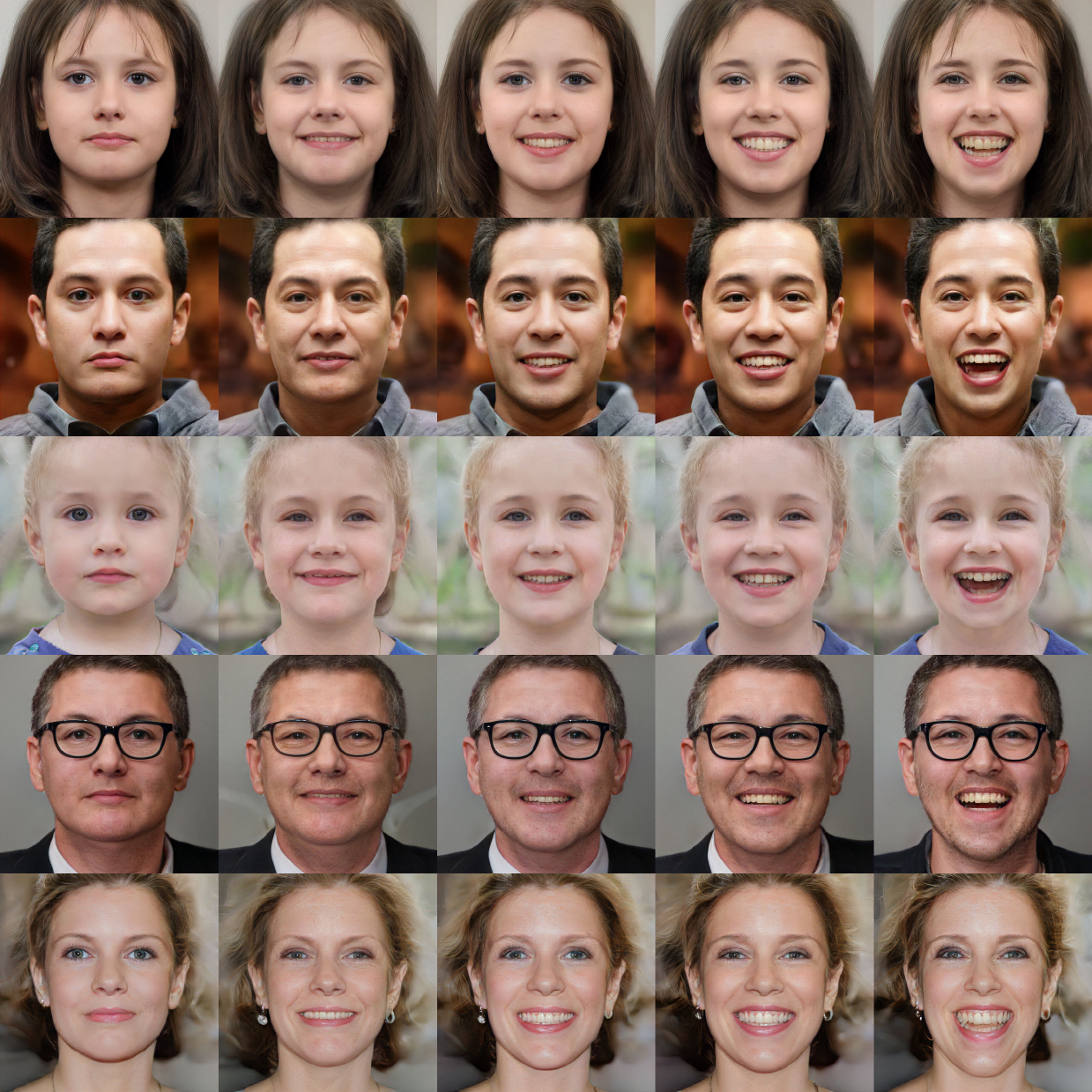} &
    \includegraphics[width=0.47\linewidth]{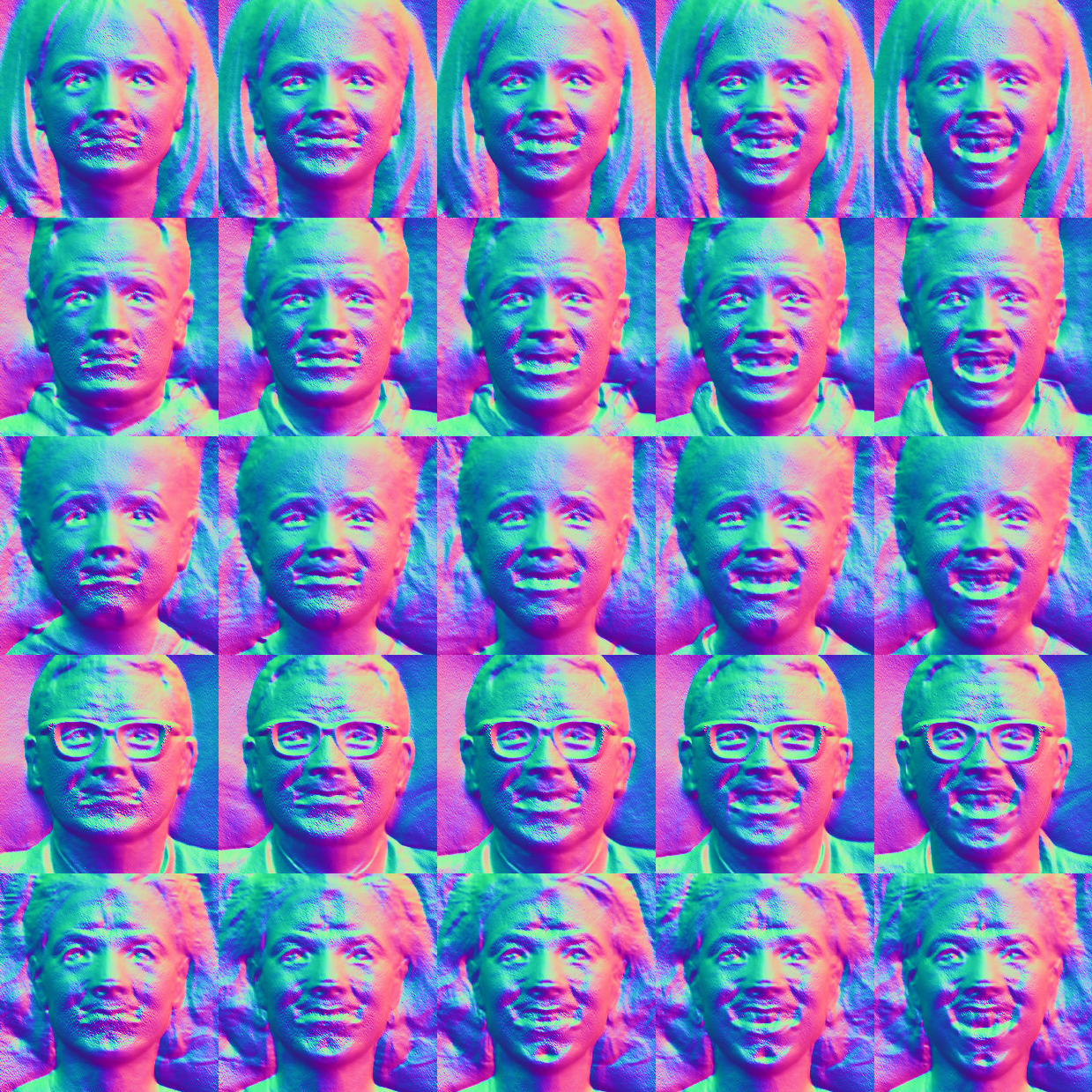} \\
    (a) Varying expressions & (b) Corresponding normal maps for (a)\\
    \includegraphics[width=0.47\linewidth]{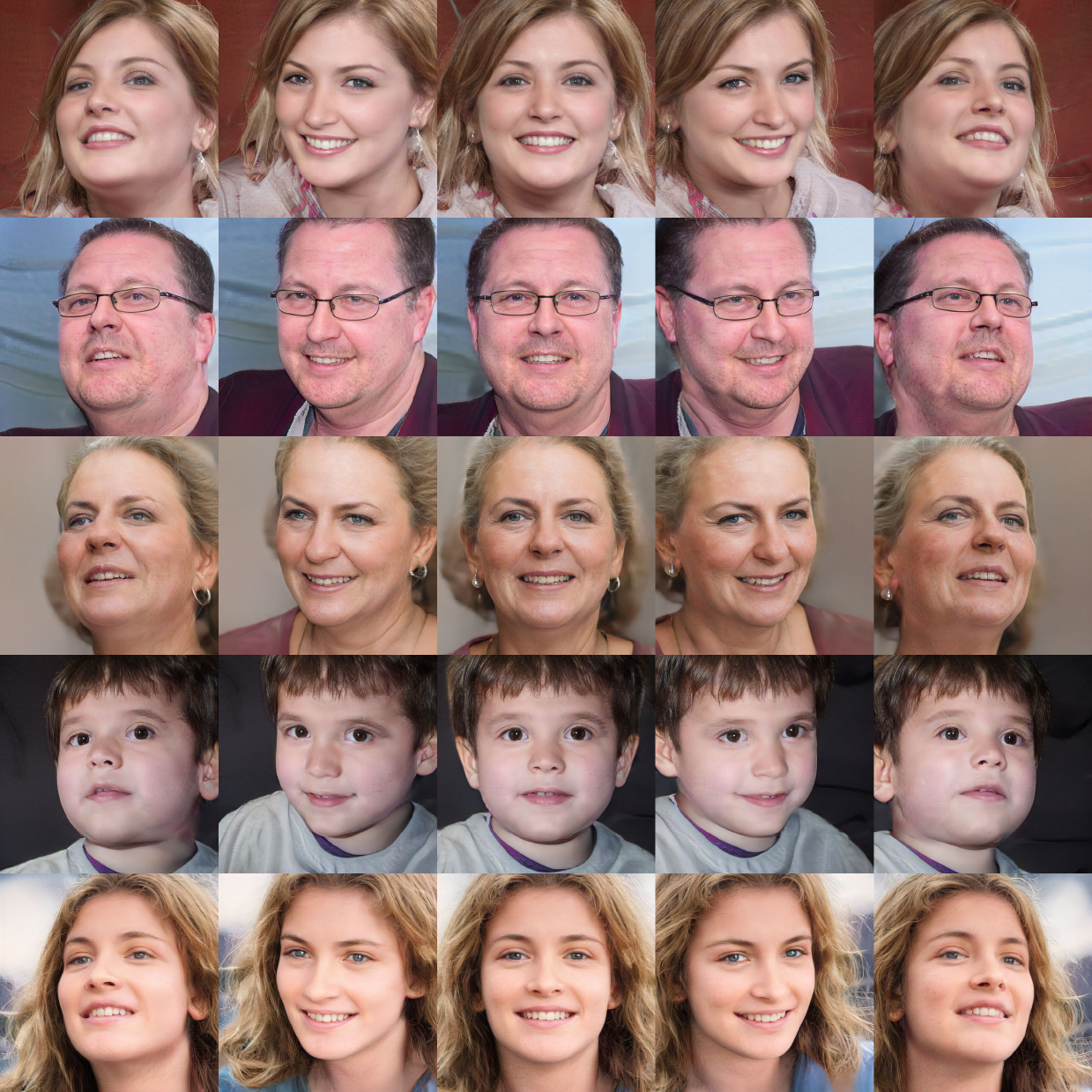} &
    \includegraphics[width=0.47\linewidth]{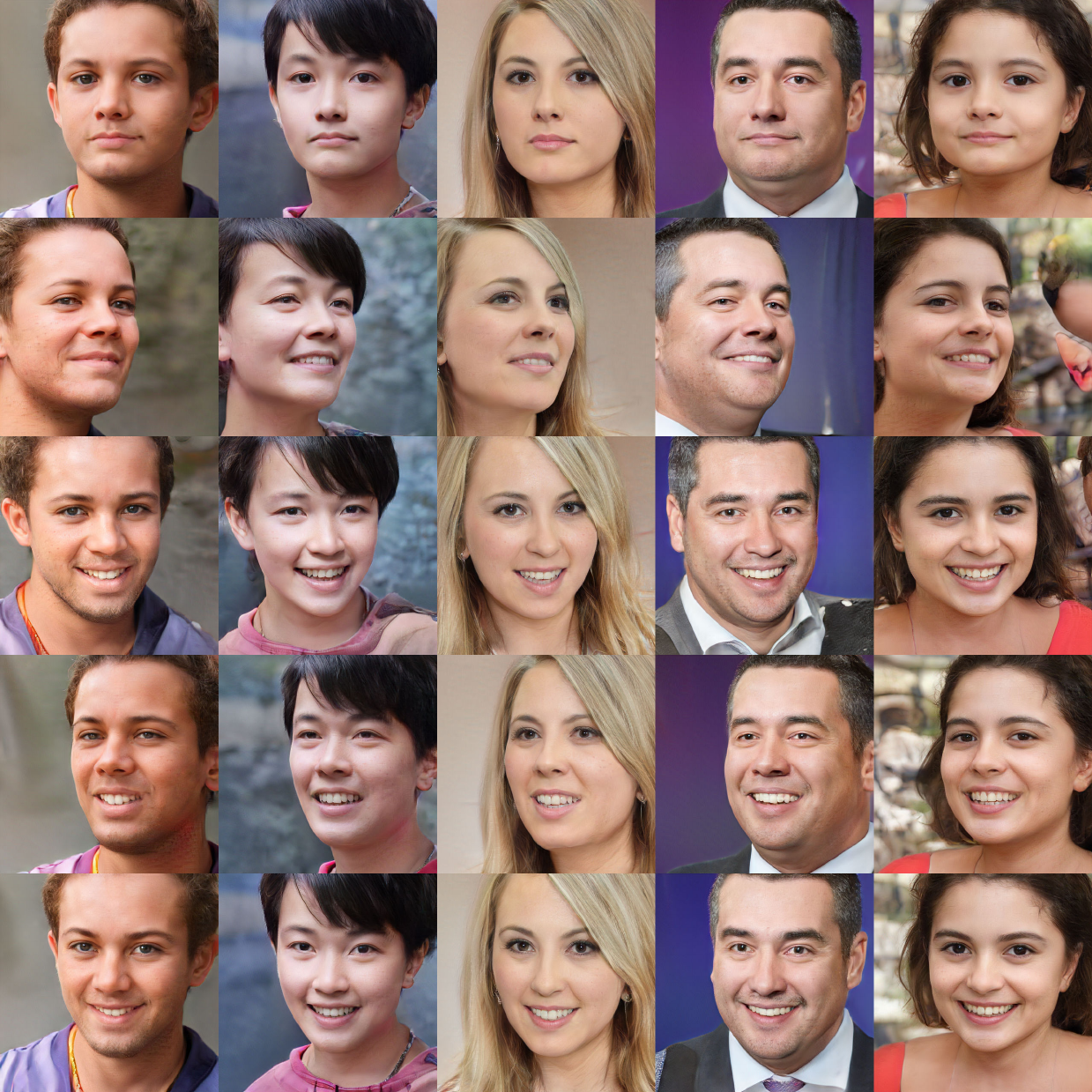} \\
    (c) Varying poses & (d) Varying identities\\
\end{tabular}
}
}
\caption{The controllable synthesis results. Face images generated by the proposed cGOF++. The variations in the expression, pose and identity of the faces are highly disentangled and can be precisely controlled.}
\label{Fig: controlled attributes}
\end{figure*}

\noindent \textbf{Datasets.}
We train our model on the {Flickr-Faces-HQ (FFHQ)}~\cite{karras2019style} dataset, which consists of $70$k high-quality face images with considerable variation in terms of age, ethnicity, facial shape, expressions, poses, and lighting conditions. We realign and crop the images to $512\times512$ following EG3D~\cite{eg3d}. For cross-domain evaluation, we also calculate Frechet Inception Distance (FID) with CelebA-HQ~\cite{lee2020maskgan}, another high-quality face image dataset.

\noindent \textbf{Metrics.}
\noindent {\it \label{metrics_ds}\textbf{(1) Disentanglement Score (DS)}.}
Following \cite{deng2020disentangled}, we evaluate 3D controllability using a 3DMM Disentanglement Score.
To measure the DS on one disentangled property $\noise_i$ (\eg shape, expression or pose), we randomly sample a set of $\noise_i$ from a normal distribution while keeping other codes fixed $\{\noise_j\}, j \neq i$, and render images.
We re-estimate the 3DMM parameters $\hat{\noise}_i$ and $\{\hat{\noise}_j\}$ from these rendered images and compute
$DS_i = \prod_{j,j \neq i} \sigma_{\hat{\noise}_i}/\sigma_{\hat{\noise}_j}$,
where $\sigma_{\hat{\noise}_i}$ and $\sigma_{\hat{\noise}_j}$ denote variance of the reconstructed parameters of the measured property $\hat{\noise}_i$ and the rest $\hat{\noise}_j$, respectively.
A higher DS score indicates that the model achieves more precise controllability on that property while keeping the others unchanged.

\noindent {\it \textbf{(2) Chamfer Distance (CD)}.}
We report the bi-diretional Chamfer Distance between the generated face surface and the input 3DMM to measure how closely the generated 3D face follows the condition.

\noindent {\it \textbf{(3) Landmark Distance (LD).}}
To evaluate the fine-grained controllability of our model, we detect the 3D landmarks in the generated images using \cite{deng2019accurate} and compute the Euclidean distance between the detected landmarks and the ones extracted from the input 3DMM vertices.

\noindent {\it \textbf{(4) Landmark Correlation (LC).}}
We measure the precision of expression control, by computing the 3D landmark displacements of two faces generated with a pair of random expression codes, and reporting the correlation score between these displacements and those obtained from the two 3DMM meshes.

\noindent {\it \textbf{(5) Average Pose Distance (APD).}}
We follow \cite{ren2021pirenderer, tang20223dfaceshop} to measure the precision of pose control by computing the error between the conditioning pose and the predicted pose of the generated image.

\noindent {\it \textbf{(6) Frechet Inception Distance (FID)}.}
The quality of the generated face images is measured using Frechet Inception Distance~\cite{DBLP:journals/corr/HeuselRUNKH17} with an ImageNet-pretrained Inception-V3~\cite{Szegedy_2016_CVPR} feature extractor. We align images with $68$ landmarks~\cite{sagonas2013300w, sun2019fab, qian2019avs} to avoid domain shift.

\subsection{Qualitative Evaluation}
\noindent \textbf{Attribute Controlling Results.}
\cref{Fig: controlled attributes} shows some example images generated by our proposed cGOF++ model.
In particular, we visualize the generated faces by varying only one of the controlled properties at a time, including facial expression, head pose, and identity--combining shape and texture as in~\cite{deng2020disentangled}.
Our model is able to generate high-fidelity 3D faces with precise 3D controllability in a highly disentangled manner, even in the presence of large expression and pose variations.
In contrast, existing methods tend to produce inconsistent face images as shown in~\cref{fig:teaser}.

\begin{figure*}
\center
\setlength\tabcolsep{0pt}
{
\renewcommand{\arraystretch}{0.0}
\footnotesize
\setlength{\tabcolsep}{1pt}{
\begin{tabular}{ccc}
    \includegraphics[width=0.29\linewidth]{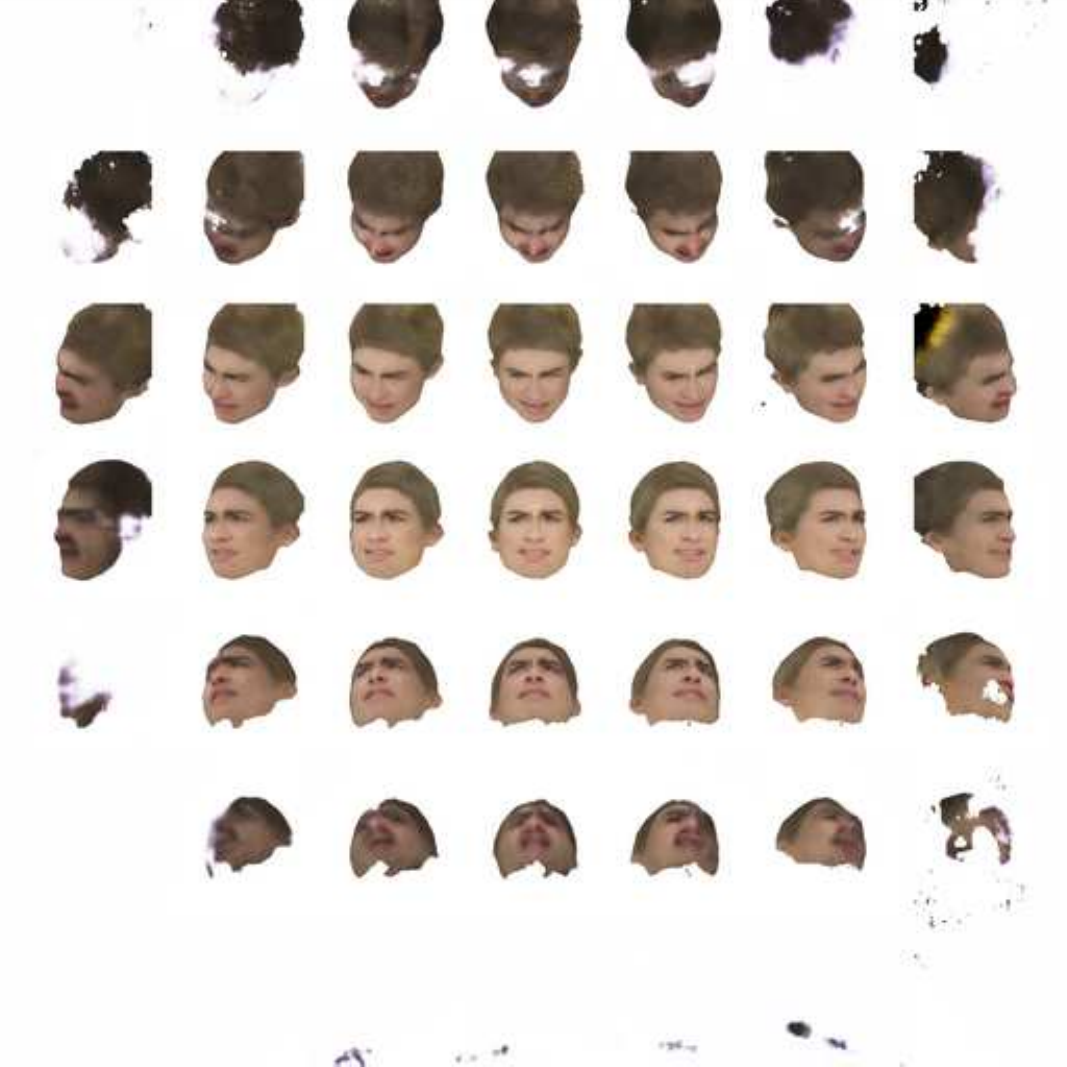} &
    \includegraphics[width=0.29\linewidth]{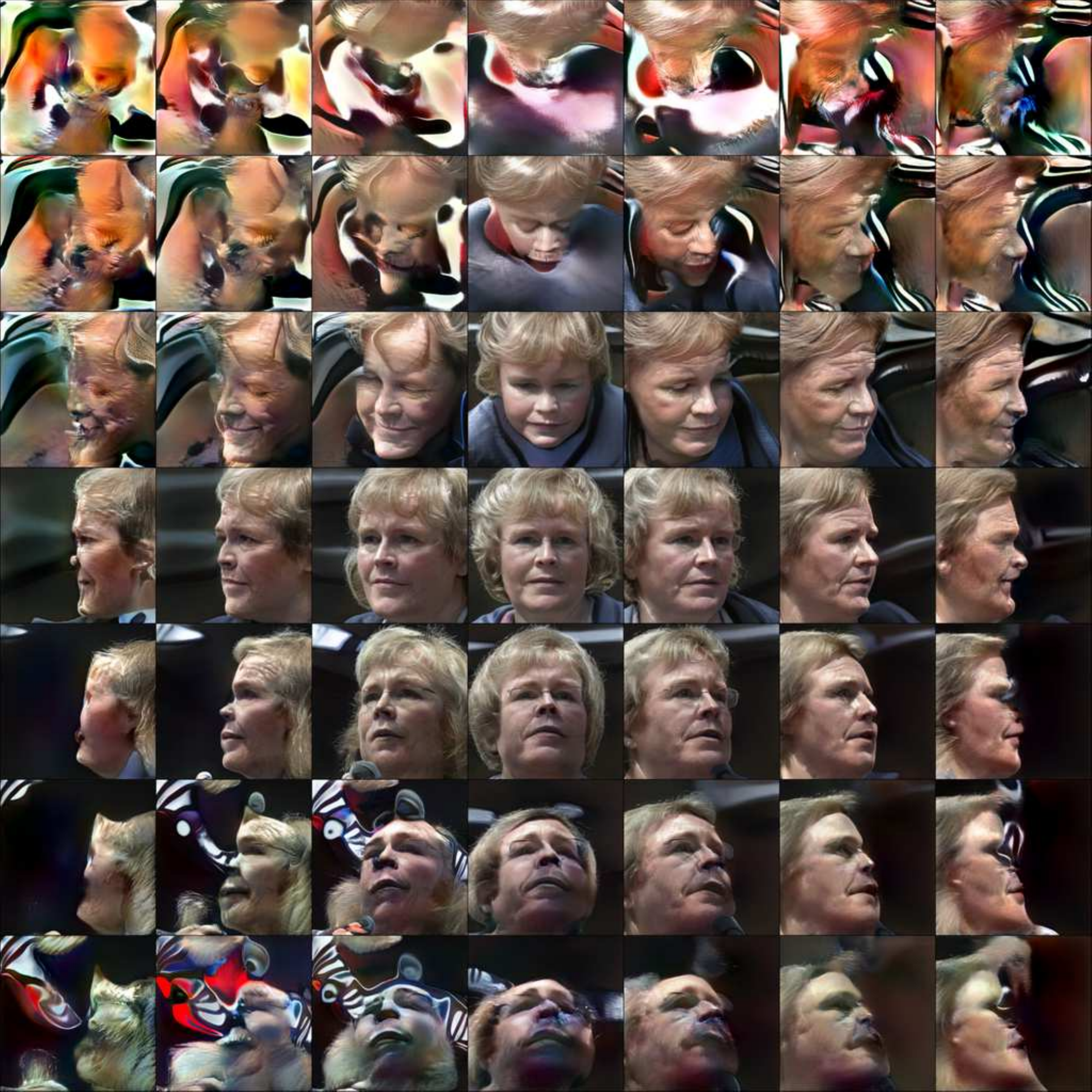} &
    \includegraphics[width=0.29\linewidth]{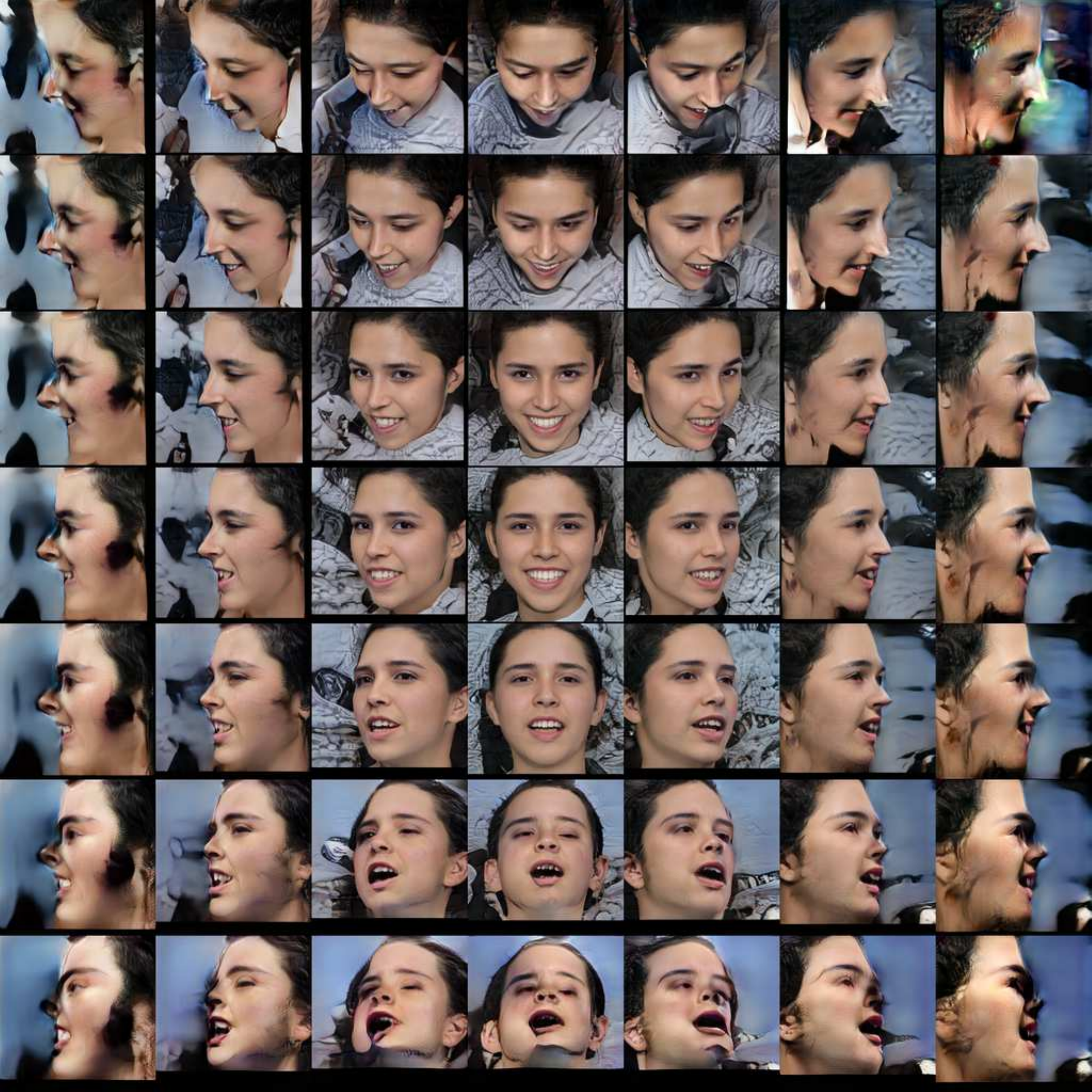} \vspace{0.1cm}\\
    (a)\label{Fig:poses_cmp:a} HeadNeRF~\cite{hong2021headnerf} & (b)\label{Fig:poses_cmp:b} GAN-Control~\cite{shoshan2021gancontrol} & (c)\label{Fig:poses_cmp:c} DiscoFaceGAN~\cite{deng2020disentangled} 
    \vspace{0.1cm}\\
    \includegraphics[width=0.29\linewidth]{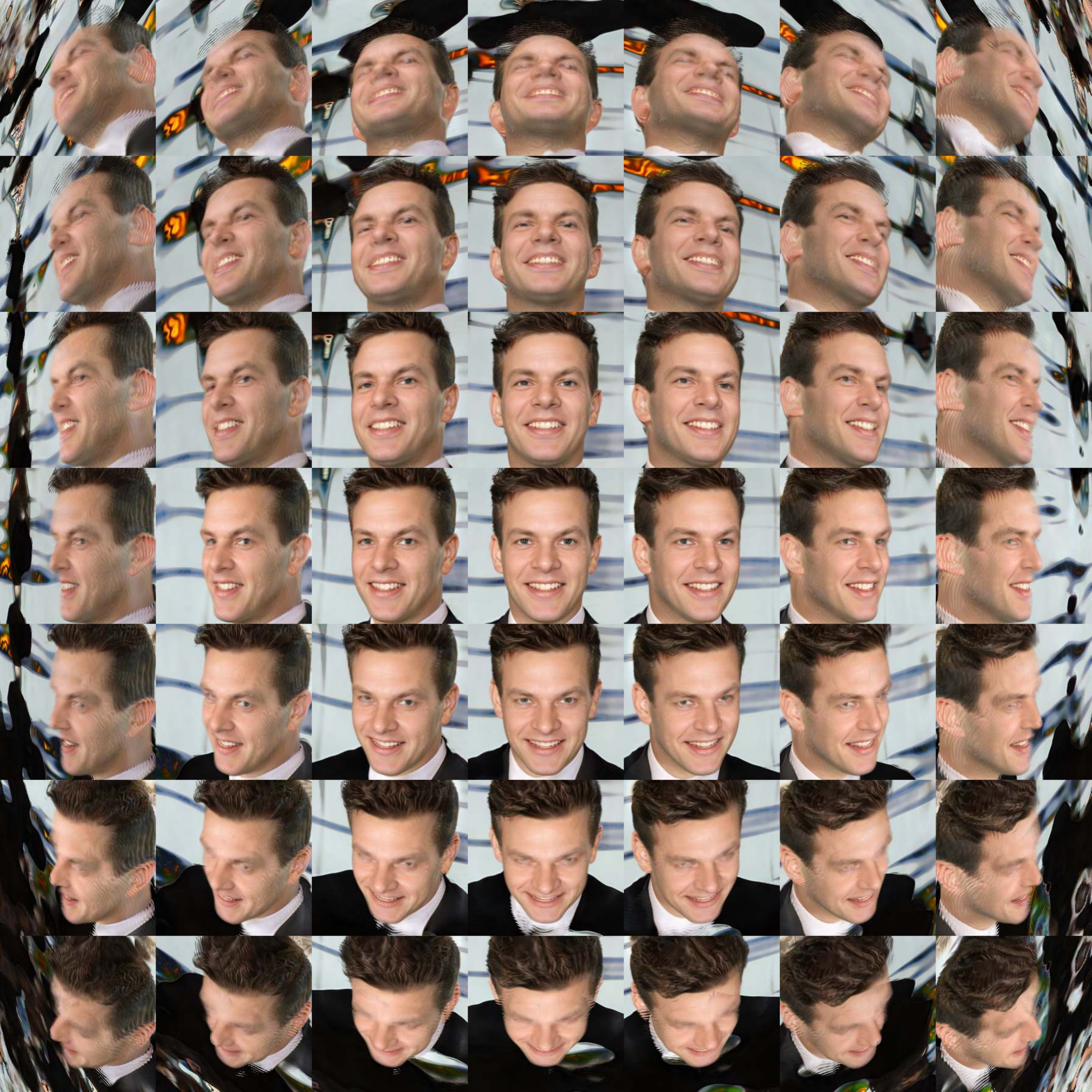} &
    \includegraphics[width=0.29\linewidth]{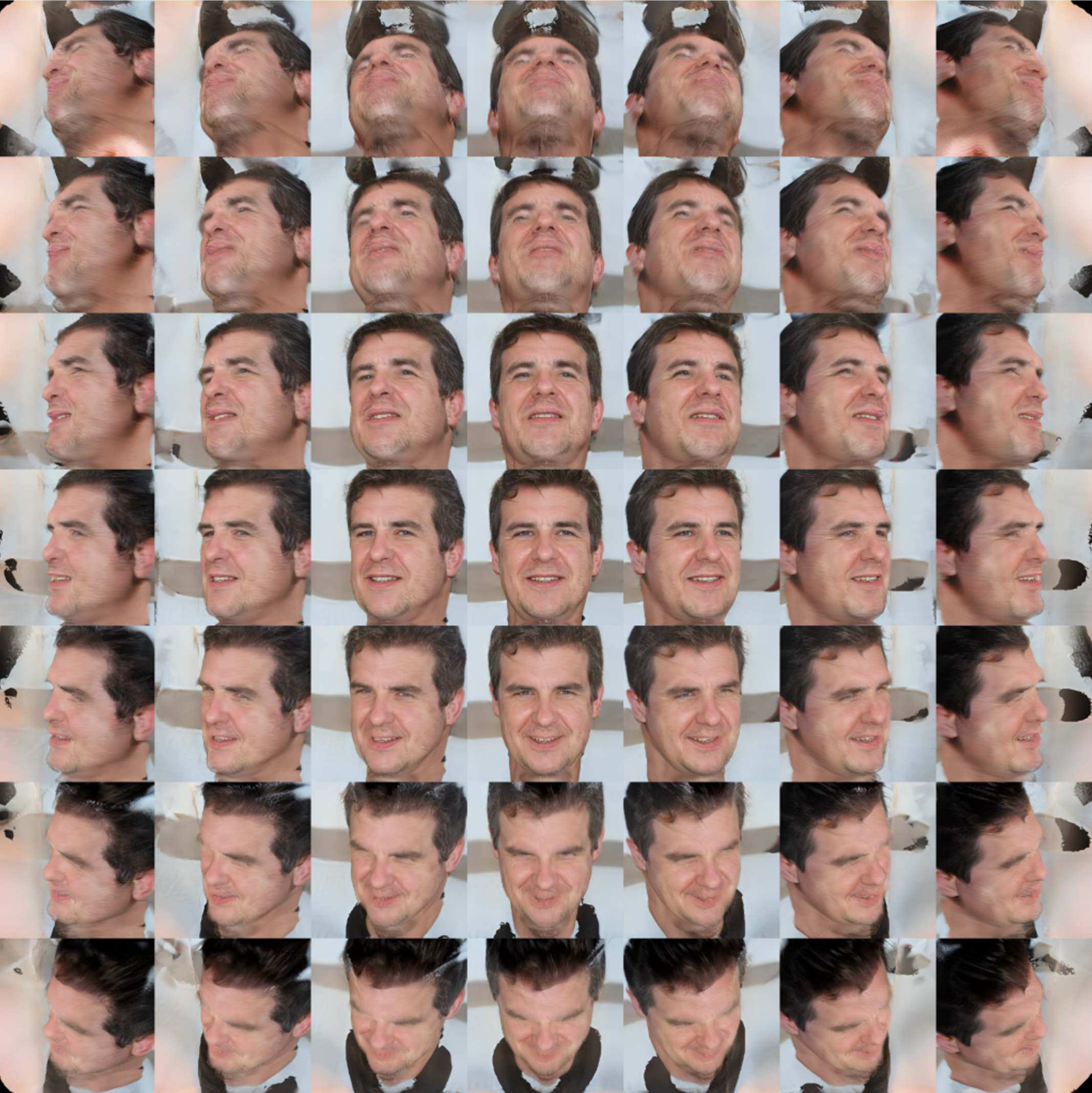} &
    \includegraphics[width=0.29\linewidth]{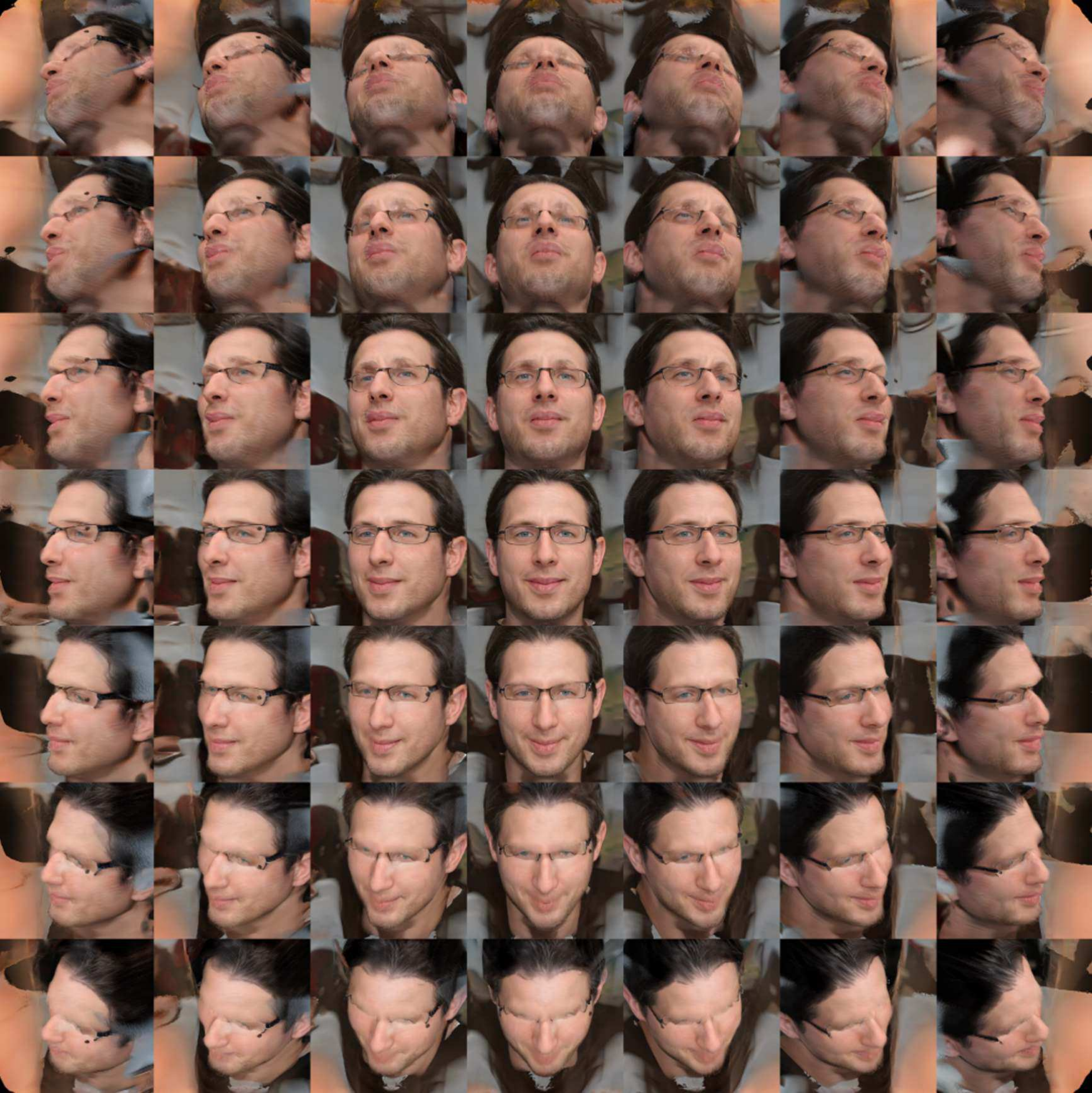} \vspace{0.1cm} \\

    (d)\label{Fig:poses_cmp:d} AniFaceGan~\cite{yue2022anifacegan} & (e)\label{Fig:poses_cmp:e} 3DFaceShop~\cite{tang20223dfaceshop} & (f)\label{Fig:poses_cmp:f} 3DFaceShop~\cite{tang20223dfaceshop} \vspace{0.1cm}\\

    \includegraphics[width=0.29\linewidth]{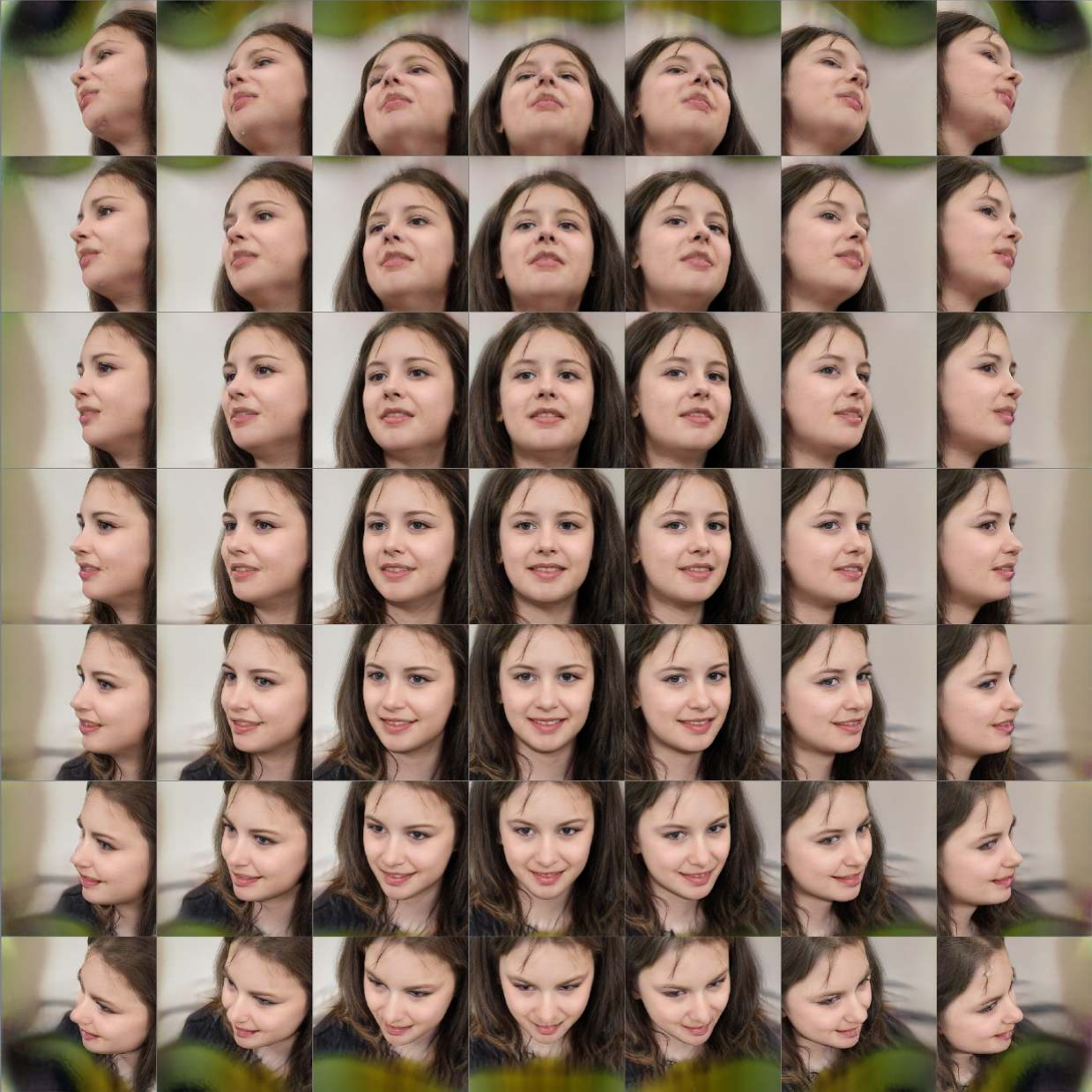} &
    \includegraphics[width=0.29\linewidth]{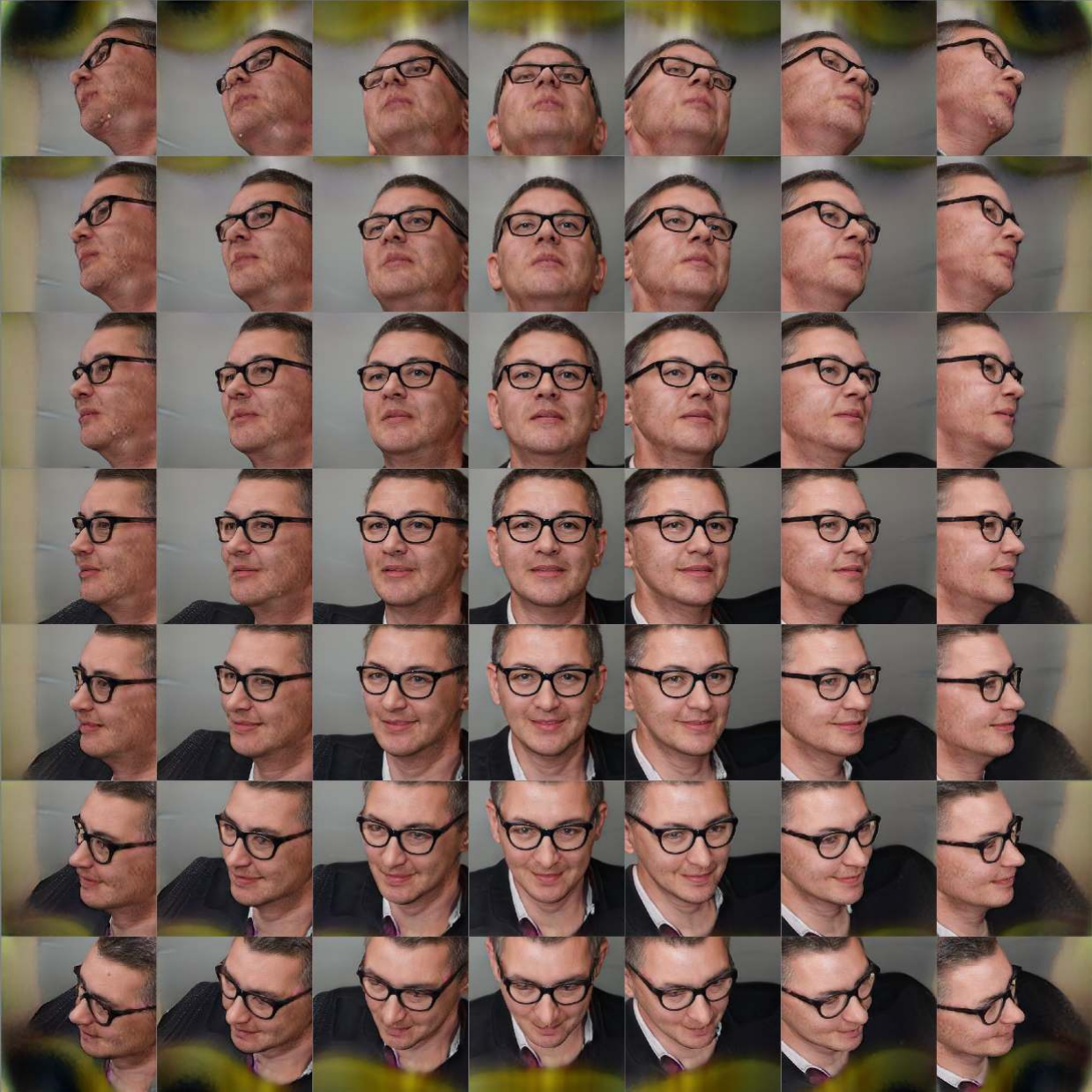} &
    \includegraphics[width=0.29\linewidth]{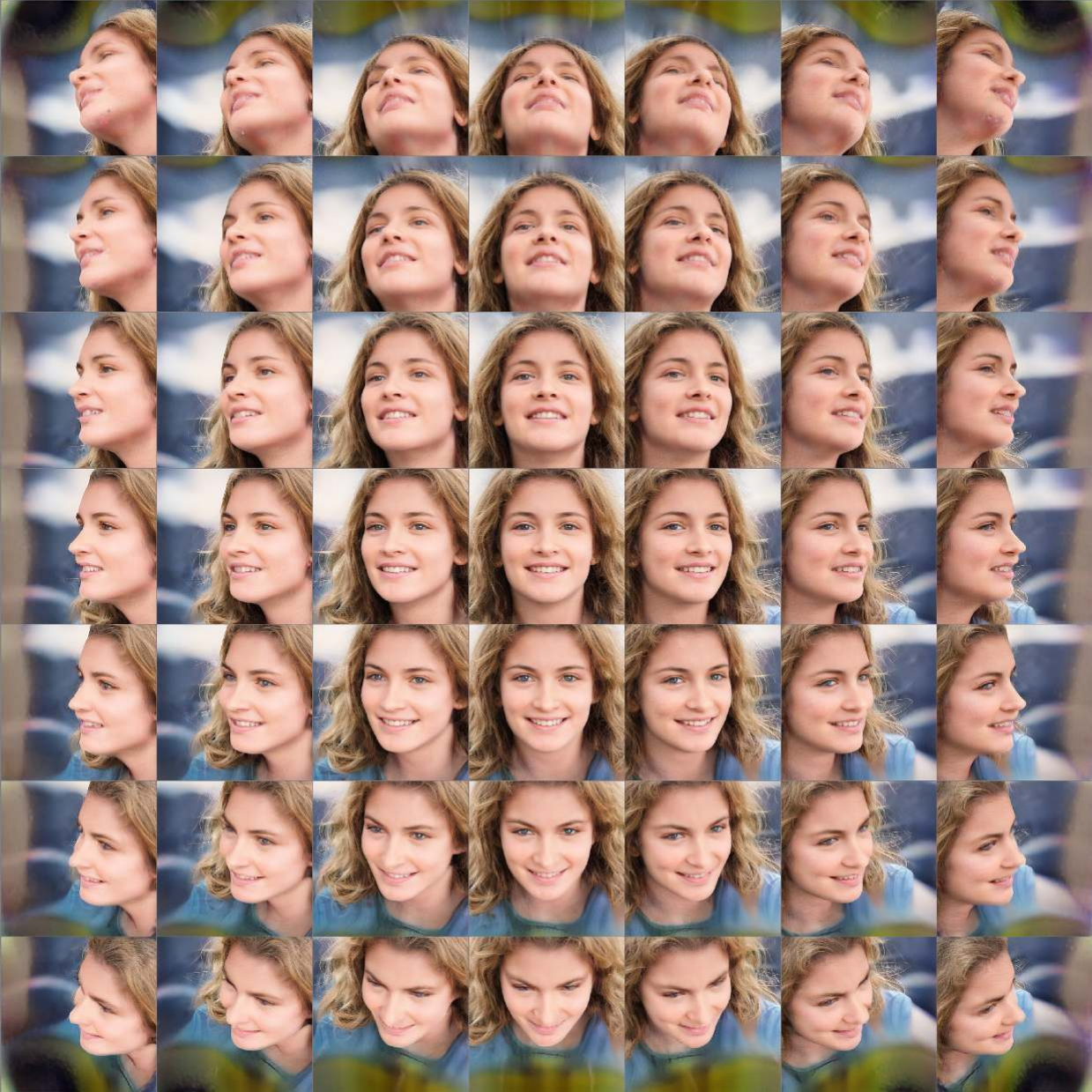} \vspace{0.1cm}\\
    
    (g)\label{Fig:poses_cmp:g} cGOF++ (Ours) & (h)\label{Fig:poses_cmp:h} cGOF++ (Ours) & (i)\label{Fig:poses_cmp:i} cGOF++ (Ours)\\
\end{tabular}}
}
\caption{
Our proposed cGOF++ surpasses existing methods in generating images with varying \textbf{poses}. Prior approaches struggle to produce satisfactory face images as the camera pose becomes larger, as shown in sub-figures (a) - (c). More recent counterparts, such as AniFaceGAN~\cite{yue2022anifacegan} and 3DFaceShop~\cite{tang20223dfaceshop}, exhibit improved performance in generating wild-pose images, as depicted in sub-figures (d)-(f). Nevertheless, the utilisation of manifolds in AniFaceGAN~\cite{yue2022anifacegan} leads to \emph{streaky artefacts} in large camera poses, while 3DFaceShop~\cite{tang20223dfaceshop} results in \emph{blurry images} when faced with challenging camera poses. Furthermore, the presence of accessories, such as glasses, exacerbates the difficulties faced by these methods.
In contrast, our method, showcased in sub-figures (g)-(i), excels at producing exquisite 3D consistent face images, even in extremely large poses, while effectively handling accessories.
}
\label{Fig:poses_cmp}
\end{figure*}

\noindent \textbf{Comparison on Pose Variations.}
We present large pose results for \headnerfc, \ganctrlc, and \discoc in \cref{Fig:poses_cmp} (a) - (c) respectively.
Previous methods fail to generate plausible face images when the camera pose gets larger (\eg $> 60^\circ$), including NeRF-based methods like \headnerfc.
More recent counterparts, such as AniFaceGAN~\cite{yue2022anifacegan} and 3DFaceShop~\cite{tang20223dfaceshop}, exhibit improved performance in generating wild-pose images, as depicted in \cref{Fig:poses_cmp} (d) - (f). Nevertheless, the utilisation of manifolds in AniFaceGAN~\cite{yue2022anifacegan} leads to \emph{streaky artefacts} in large camera poses, while 3DFaceShop~\cite{tang20223dfaceshop} results in \emph{blurry images} when faced with challenging camera poses. Furthermore, the presence of accessories, such as glasses, exacerbates the difficulties faced by these methods.
In contrast, our method, showcased in \cref{Fig:poses_cmp} (g) - (i), excels at producing exquisite 3D consistent face images, even in extremely large poses, while effectively handling accessories.

\begin{figure*}
\center
\setlength\tabcolsep{0pt}
{
\renewcommand{\arraystretch}{0.0}
\footnotesize

\begin{tabular}{cc}
    \includegraphics[width=0.46\linewidth]{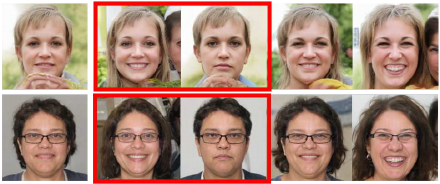} &
    \includegraphics[width=0.46\linewidth]{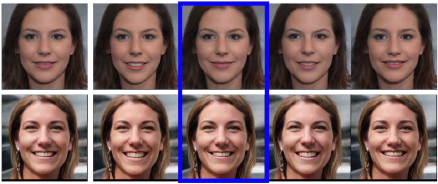}\\
    (a) GANControl~\cite{shoshan2021gancontrol} & (b) DiscoFaceGAN~\cite{deng2020disentangled} \\
    \includegraphics[width=0.46\linewidth]{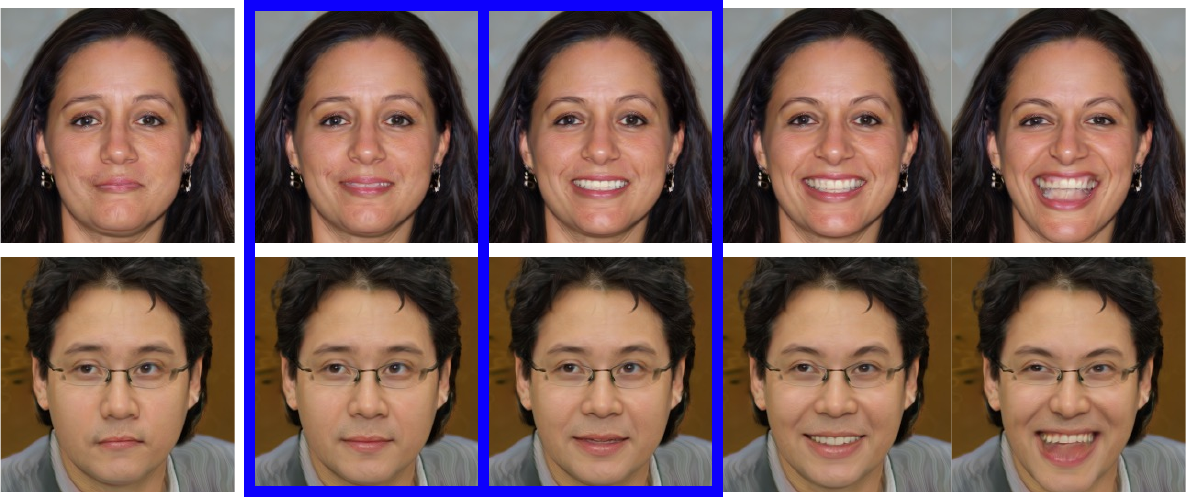} &
    \includegraphics[width=0.46\linewidth]{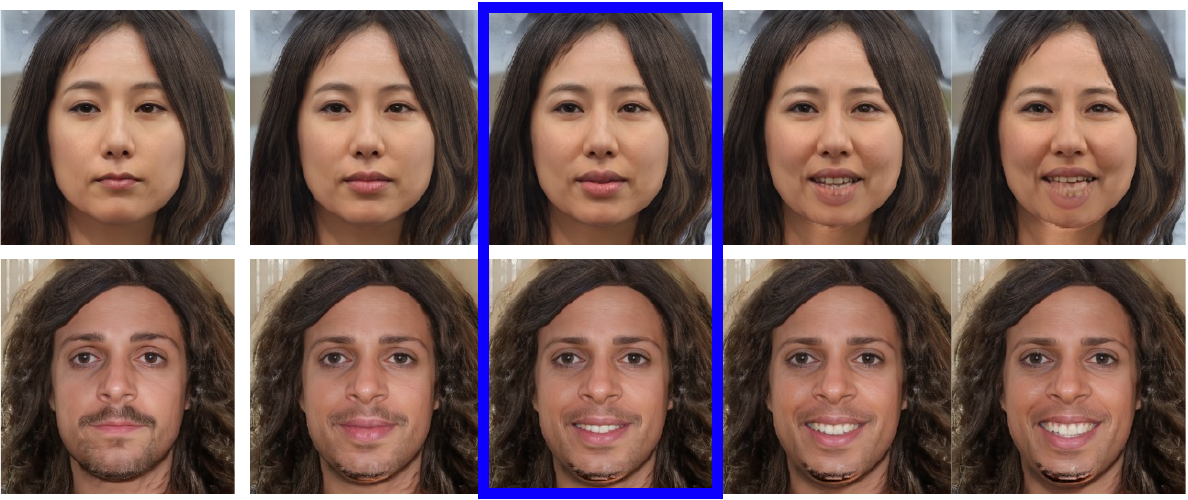}\\
    (c) AniFaceGAN~\cite{yue2022anifacegan} & (d) 3DFaceShop~\cite{tang20223dfaceshop} \\
\multicolumn{2}{c}{}
    \includegraphics[width=0.94\linewidth]{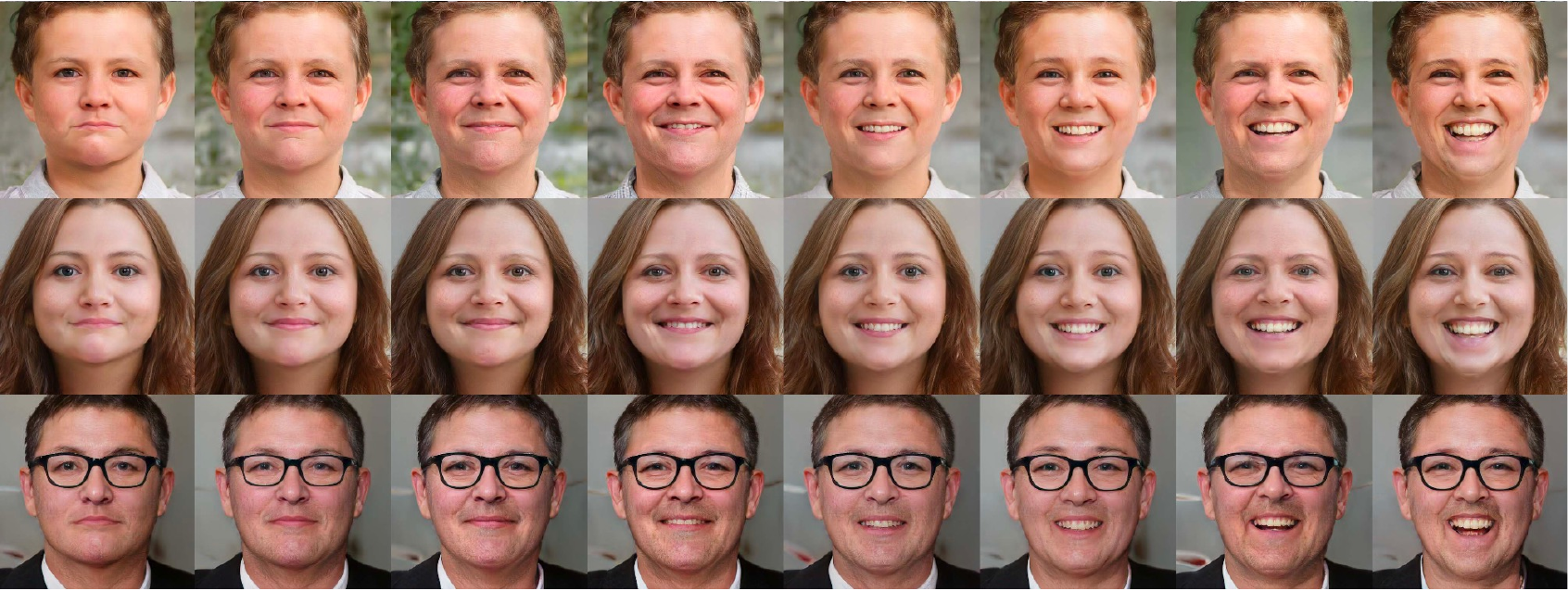}\\
\multicolumn{2}{c}{(e) cGOF (Ours)}

\end{tabular}
}
\caption{Images generated by our proposed cGOF with different \textbf{expressions}, compared with existing methods. In each sub-figure (a) - (e), each \textbf{row} shares \textbf{the same identity code}, while each \textbf{column} has \textbf{the same expression code}. We use \textbf{Red} boxes to highlight the identity inconsistency, and use \textbf{Blue} boxes to highlight the expression inconsistency.}
\label{Fig:exp_cmp}
\end{figure*}

\noindent \textbf{Comparison on Expression Variations.}
We present more results on the expression control in \cref{Fig:exp_cmp}, comparing our method against \emph{four} state-of-the-art controllable face synthesis methods, one attribute-guided \ganctrlc, and the other 3DMM-guided \discoc, AniFaceGAN~\cite{yue2022anifacegan}, 3DFaceShop~\cite{tang20223dfaceshop}.
For each sub-figure, each row corresponds to the same person, and each column corresponds to the same expression.

\begin{figure}[t!]
\begin{center}
\resizebox{1.0\linewidth}{!}{
\includegraphics[width=1\linewidth]{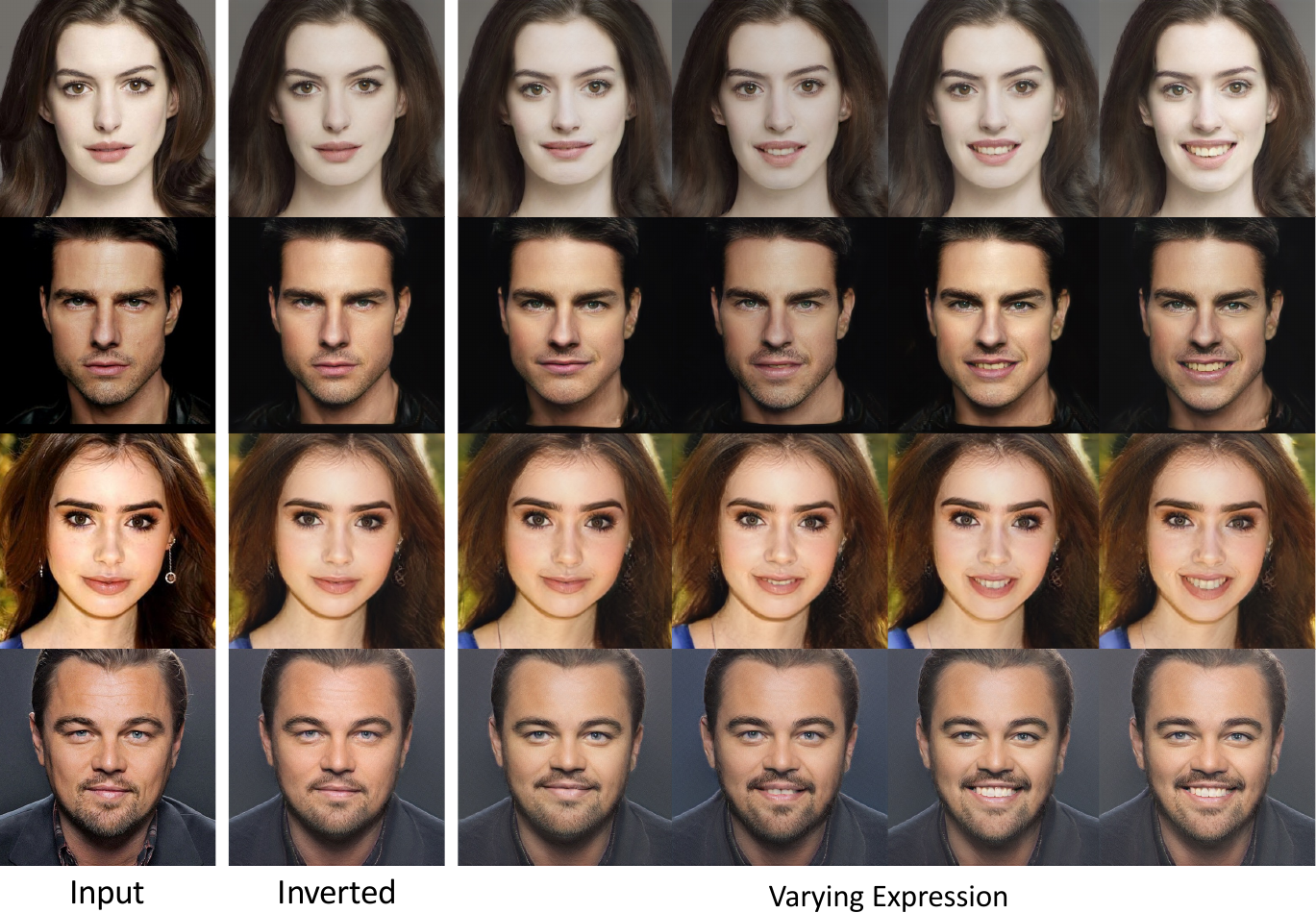}
}
\end{center}
\caption{Inversion and editing results. We present faithful inversion results and vivid facial expression control results.
}
\label{fig:inversion_editing_results}
\vspace{-5pt}
\end{figure}

In \cref{Fig:exp_cmp} (a), we can see that \ganctrl fails to preserve the identity as well as other factors of the face image (\eg background) when changing only the facial expression code.
A few examples are highlighted in red.
Moreover, we observe that with the original range of the expression parameters, the model results in only a small variation of expressions, whereas with increased perturbations, it leads to much more significant shape and identity inconsistencies.
We also notice the ``smiling'' attribute tends to strongly correlate with the ``female'' and ``long-hair'' attributes.
In \cref{Fig:exp_cmp} (b), we can see that the \disco fails to impose consistent expression control over the faces.
The blue boxes highlight a few examples, where the same expression code produces different expressions in different faces.
As depicted in Figure \ref{Fig:exp_cmp} (c) and (d), AniFaceGAN~\cite{yue2022anifacegan} and 3DFaceShop~\cite{tang20223dfaceshop} encounter challenges in preserving expression consistency. These methods struggle to ensure that the generated facial expression remains consistent even when provided with the same expression code.
In \cref{Fig:exp_cmp} (e), we show that our model generates compelling photo-realistic face images with highly consistent, precise expression control.
In each row, only expressions change while other properties remain unchanged, such as identity (shape and texture), hair, and background.

\begin{table*}[t!]
\caption{\label{tab:ds_cmp}
DS and FID results of state-of-the-art controllable face synthesis methods}

\centering
\begin{tabular}{llcccccccc}
\hline
\multicolumn{1}{c}{Method Name}             &Venue          & $DS_\text{s}$ $\uparrow$ & $DS_\text{e}$ $\uparrow$ & $DS_\text{p}$ $\uparrow$   & LD $\downarrow$ & LC (\%) $\uparrow$ & APD (\%) $\downarrow$ &FID (CelebAHQ)$\downarrow$ &FID (FFHQ)$\downarrow$ \\
\hline
PIE~\cite{tewari2020pie}                    &TOG'20         &1.66	            &15.24	            &2.65                 & -              & -     & -       &82.15            & 80.64\\
StyleRig~\cite{tewari2020stylerig}          &CVPR'20        &1.64	            &13.03	            &2.01                 & -              & -     & -       &109.07            & 109.94\\
DiscoFaceGAN~\cite{deng2020disentangled}    &CVPR'20        &5.97	            &\underline{15.70}  &5.23                 & 2.82            & \underline{70.81}     & 0.16       &160.24            & 116.43 \\
E4E~\cite{e4e}                              &SIGGRAPH'21    &1.91	            &8.66	            &7.08                 & -               & -      & -      &87.59            & 68.46 \\
Gan-Control~\cite{shoshan2021gancontrol}    &ICCV'21        &\underline{7.07}	&7.51	            &9.33                 & -               & -      & 0.22      &80.52            & 50.66 \\
HeadNeRF~\cite{hong2021headnerf}            &CVPR'22        &6.39	            &5.99	            &10.26                & -               & -      & 0.15      &139.09        & 157.16  \\
AniFaceGAN~\cite{yue2022anifacegan}         &NeurIPS'22     &  6.44	            &14.28	            &\underline{15.62}    & -               & -      & -     & 74.47       & 56.28  \\
3DFaceShop~\cite{tang20223dfaceshop}        &TVCG'23       & 5.86	            &11.81	            &12.97                & \underline{2.79}            & 68.66      & \underline{0.10}     & \underline{60.20}       & \underline{41.58}  \\
Ours                                        &-              &\textbf{10.22}     &\textbf{16.10}     &\textbf{19.63}       & \textbf{1.81}   & \textbf{82.17}  & \textbf{0.08}   &\textbf{53.14}     &\textbf{18.55}                   \\

\hline
\end{tabular}

\end{table*}

\begin{table*}[t!]
    \caption{\label{tab:user_study}
        User Study on the Disentangle Performance and Image Quality.}
    \centering
    \footnotesize
    \resizebox{1.0\linewidth}{!}{
    \begin{tabular}{lccccc}
    \hline
        ~ & DiscoFaceGAN~\cite{deng2020disentangled} & GAN-Control~\cite{shoshan2021gancontrol} & Pi-GAN~\cite{pigan} + $\mathcal{L}_\text{recon}$ & HeadNeRF~\cite{hong2021headnerf} & cGOF (Ours) \\ \hline
        Id & 3.26 / 54.8 / 2.63\% & 2.55 / 69.0 / 18.42\% & 3.21 / 55.8 / 7.89\% & 4.16 / 36.8 / 13.16\% & \textbf{1.82 / 83.6 / 57.89\%} \\
        Exp & 2.45 / 71.0 / 28.95\% & 3.34 / 53.2 / 10.53\% & 3.47/ 50.6 / 2.63\% & 3.63 / 47.4 / 15.79\% & \textbf{2.11 / 77.8 / 42.11\%} \\
        Pose & 2.84 / 63.2 / 18.42\% & 3.82 / 43.6 / 2.63\% & 2.21 / 75.8 / 18.42\% & 4.29 / 34.2 / 2.63\% & \textbf{1.84 / 83.2 / 57.89\%}\\
        IQ & 2.45 / 71.0 / 15.79\% & 3.34 / 53.2 / 5.26\% & 3.18 / 56.4 / 2.63\% & 4.61 / 27.8 / 5.26\% & \textbf{1.42 / 91.6 / 71.05\%} \\
        \hline
    \end{tabular}
    }
    {Each cell contains "Average Ranking / Average Score / Ranking 1st Ratio". Id: Identity, Exp: Expression, IQ: Image Quality.}
    
\end{table*}

\noindent \textbf{Inversion and Editing Results} \cref{fig:inversion_editing_results} shows the inversion and editing results as introduced in \cref{inv_edit}. In the first column are sampled in-the-wild images, we inverse the cGOF++ to obtain the inversion results in the second column. The following four columns are expression edited results, where we see the facial expression varies from smiling to grinning.

\subsection{Quantitative Evaluation}
We compare our method to several state-of-the-art methods.
E4E~\cite{e4e} and GAN-Control~\cite{shoshan2021gancontrol} are purely 2D controllable face image synthesis methods, which are based on facial attribute annotations.
DiscoFaceGAN~\cite{deng2020disentangled}, PIE~\cite{tewari2020pie} and StyleRig~\cite{tewari2020stylerig} leverage 3DMM priors but still use a 2D image-based StyleGAN generator.
HeadNeRF~\cite{hong2021headnerf} uses a 3D \nerf representation as well as 3DMM but is trained with a reconstruction loss using annotated multi-view datasets, whereas our method is a 3D generative model trained on \emph{unannotated single-view} images only.

\begin{table*}[t!]
\caption{\label{tab:ablation}
Ablation Study
}
\centering
\begin{tabular}{clcccccccc}
\hline
\multicolumn{1}{l}{Index} & Loss & CD $\downarrow$  & LD $\downarrow$   & LC (\%) $\uparrow$   & $DS_\text{s}$ $\uparrow$ & $DS_\text{e}$ $\uparrow$ & $DS_\text{p}$ $\uparrow$ & FID (128) $\downarrow$  \\
\hline
1 &$\mathcal{L}_\text{gan}$
&2.47	            &2.50	            &0.88	            &7.86       &6.84       &14.93       &\textbf{24.38}         \\
2 &~+ $\mathcal{L}_\text{recon}$
&2.35	            &2.10	            &20.84	            &8.16       &9.84       &20.58      &25.38                  \\
3 &~+ MgS
&\textbf{0.52}	            &2.02	            &28.30	            &8.81       &10.92       &19.47   &25.30                  \\
4 &~+ $\mathcal{R}_\text{d}$
&0.91               &2.05               &31.06              &8.50       &10.42       &17.26   &26.59                  \\
5 &~+ $\mathcal{L}_\text{ldmk}$
&0.70               &1.04               &70.43              &14.45      &16.51       &21.53   &27.60                  \\
6 &~+ $\mathcal{L}_\text{warp}$
&0.74      &\textbf{1.00}               &\textbf{73.30}              &\textbf{14.57}      &\textbf{16.86}       &\textbf{21.74}   &31.47                  \\
\hline
\end{tabular}

Each row adds an additional loss term into the objective function upon all the loss terms above it. The last row is our final model. 
\end{table*}

\noindent \textbf{Control Disentanglement}
We compare the control disentanglement with state-of-the-art methods using the Disentanglement Score (see Col $3$ to $5$ of \cref{tab:ds_cmp}).
Our method achieves significantly better disentanglement and more precise control in terms of shapes, expressions, and poses, compared to existing methods.
It is worth noting that we also compare with concurrent 3D-based models~\cite{tang20223dfaceshop, yue2022anifacegan}.
3DFaceShop~\cite{tang20223dfaceshop} does not exhibit superiority over existing methods in terms of disentangled scores because it mainly acquires the model's controllability by exerting photometric supervision on the images, which is inefficient for 3D control, rather than directly on the 3D volume.
AniFaceGAN~\cite{yue2022anifacegan} achieves independent control over face shape, expression, and head pose, leading to outstanding disentangle scores. However, since there is no clear surface, the warping field in \cite{yue2022anifacegan} fails to warp the face accurately, which explains the gap between their approach and our proposed method.

Note that since PIE and StyleRig do not have open-sourced code, we compute the score on the $167$ generated images provided on the PIE project page\footnote{\url{https://vcai.mpi-inf.mpg.de/projects/PIE/}}.

\noindent \textbf{Control Accuracy}
We also conduct an analysis of our proposed method against state-of-the-art face synthesis models~\cite{deng2020disentangled, tang20223dfaceshop} that utilize facial landmarks for control, in terms of the Landmark Distance (LD), Landmark Correlation (LC), and Average Pose Distance (APD)~\cite{ren2021pirenderer}, and the results are presented in Col $6$ to $8$ of \Cref{tab:ds_cmp}.
Our proposed method outperforms the prior methods across all the LD, LC and APD metrics. This suggests that our method also achieves highly accurate control over shape, expression, and poses.

\noindent \textbf{Image Quality}
The FID scores are reported in Col $9$ and $10$ of \cref{tab:ds_cmp}.
We perform the cross-dataset FID evaluation to comprehensively assess and compare the image quality of different generative models. 
While we evaluate FID scores for all compared methods on FFHQ, it's important to note that different alignments and crop schemes may have an impact on the final results.
To reduce the impact of these factors, we report FID results on both datasets to build a comprehensive benchmark to evaluate the image quality of existing methods.

The images are generated with a relatively small viewpoint and expression variation akin to the real image dataset, where most existing methods can still produce reasonable images.
These results do not take into account the failure cases with large pose variations, as shown in \cref{Fig:exp_cmp} (a) - (c).
The results show that our model outperforms existing methods not only in terms of disentanglement score, but also in the image quality.

\subsection{\label{user_study}User Study}

We conduct a user study to add to the comparison. We follow the experiment setting as in \cref{tab:ds_cmp}. Specifically, we provide the control results of five methods in random order, and ask a total of $38$ users to rank the results according to their image quality and the controlling effects.

\cref{tab:user_study} summarizes the results, where the "Average Ranking / Average Score / Ranking 1st Ratio" are reported for each method with respect to concerning factors, \ie identity, expression, pose.
Our model achieves the highest average ranking and scores for all aspects, indicating our model produces more perceptually compelling results and achieves better 3D controllability than other counterparts.

\begin{figure}[t!]
\begin{center}
\resizebox{1\linewidth}{!}{
\includegraphics[width=1\linewidth]{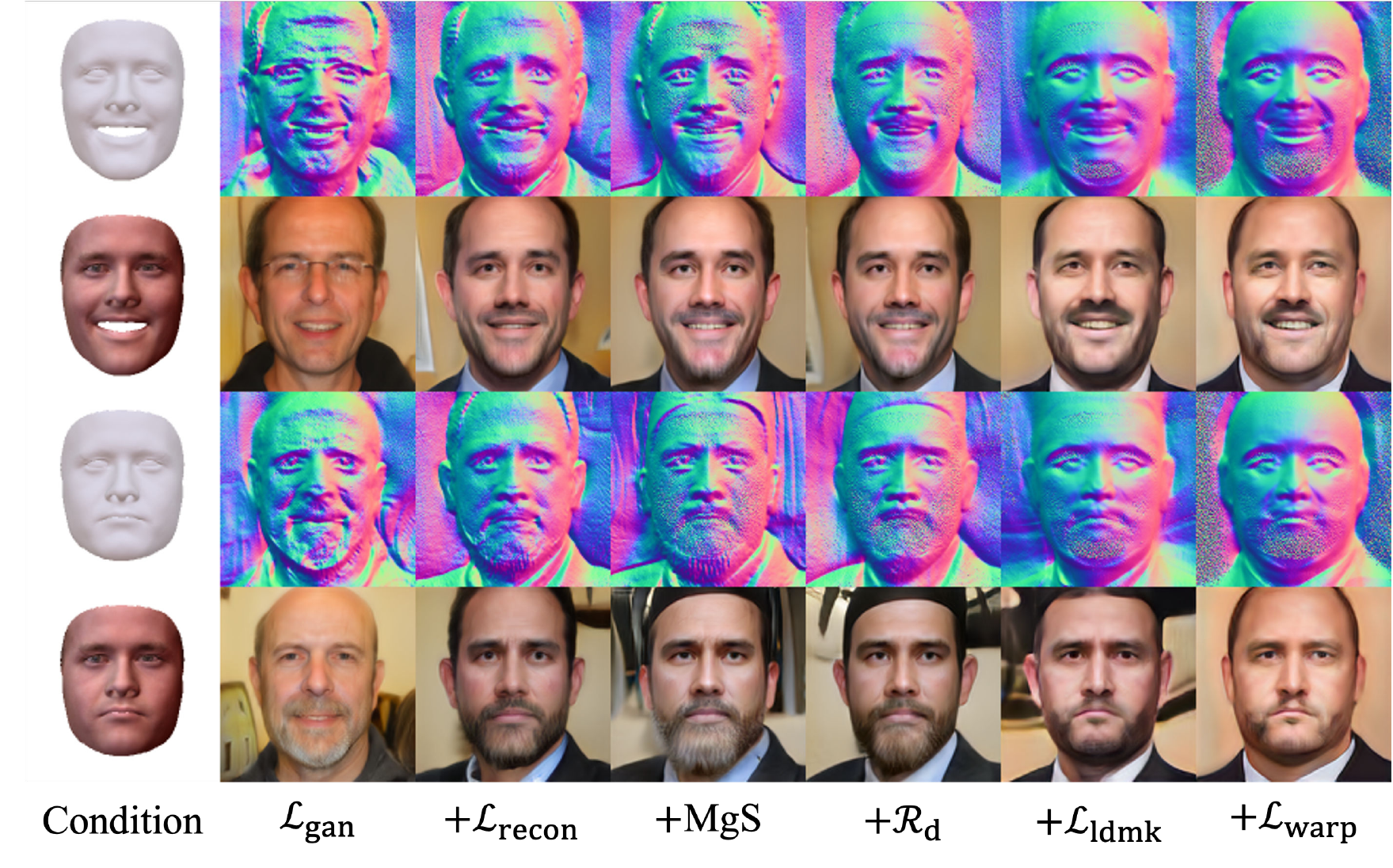}
}
\end{center}
\caption{Ablation Study. We analyze the effects of the individual components and loss functions.
With the components gradually enabled, the model is able to generate 3D realistic faces precisely following the condition 3DMM.
Refer to \cref{subsec_ablation} for more details.
}
\label{fig:ablation}
\vspace{-10pt}
\end{figure}

\subsection{\label{subsec_ablation}Ablation Study}

We conduct ablation studies to validate the effectiveness of each component in our model by adding them one by one, and evaluate the generated images with the four metrics specified above.
\cref{tab:ablation} and \cref{fig:ablation} illustrate the studies and show that all the components improve the precision of 3D controllable image generation.
In particular, the mesh-guided volume sampler (MgS) significantly improves the effectiveness of the 3DMM condition (denoted as `+ MgS' in~\cref{tab:ablation}).
The 3D landmark loss $\mathcal{L}_\text{ldmk}$ (denoted as `+ $\mathcal{L}_\text{ldmk}$') and the volume warping loss $\mathcal{L}_\text{warp}$ (denoted as `+ $\mathcal{L}_\text{warp}$') further enhance the control over fine-grained geometric details.
We also report FIDs of the outputs of the cGOF++ with the image resolution of $128 \times 128$.
Overall, although the geometric regularizers lead to slightly degraded image quality (indicated by increased FIDs), they improve the 3D controllability by large margins (indicated by the improvement in other metrics).

\section{\label{sec_limitation}Limitations.}
Since we focus on the control of facial identity and expression, the proposed method still lacks precise texture and illumination control over the generated face images.
Also, the proposed method relies on an off-the-shelf 3DMM predictor, which results in the inefficiency of our model in modelling non-neutral expressions like closing eyes or frowning.

Although our work has surpassed similar methods in terms of disentangled control, there are still noticeable limitations in attribute disentanglement in some cases. The introduction of additional face identity supervision, such as a face recognition model, might alleviate the problem at hand.

\section{Conclusions}
We have presented a controllable 3D face synthesis method that can generate high-fidelity face images based on a conditional 3DMM mesh.
We propose a novel conditional Generative Occupancy Field representation (cGOF++) and a set of 3D losses that effectively impose 3D conditions directly on the generated \nerf volume, which enable much more precise 3D controllability over 3D shape, expression, pose of the generated faces than state-of-the-art 2D counterparts.
{\label{sec_societal}\bf Societal Impacts.} The proposed work might be used to generate fake facial images for imposture or to create synthesized facial images to hide the true identities of cyber criminals. Defensive algorithms on recognizing such synthesized facial images should be accordingly developed.

\ifCLASSOPTIONcompsoc
  \section*{Acknowledgments}
\else
  \section*{Acknowledgment}
\fi

This project is funded in part by National Key R\&D Program of China Project 2022ZD0161100, by the Centre for Perceptual and Interactive Intelligence (CPII) Ltd under the Innovation and Technology Commission (ITC)’s InnoHK, by General Research Fund of Hong Kong RGC Project 14204021. Hongsheng Li is a PI of CPII under the InnoHK. 

We would like to thank Eric Ryan Chan for sharing the well structured code of the wonderful project EG3D. We thank Yu Deng for providing the evaluation code of the Disentanglement Score, and Jianzhu Guo for providing the pre-trained face reconstruction model. We are also indebted to thank Xingang Pan, Han Zhou, KwanYee Lin, Jingtan Piao, and Hang Zhou for their insightful discussions.

{\small
\bibliographystyle{IEEEtran}
\bibliography{ref}
}

\ifCLASSOPTIONcaptionsoff
  \newpage
\fi

\vspace{-0.5cm}
\begin{IEEEbiography}[{\includegraphics[width=1in,height=1.25in,clip,keepaspectratio]{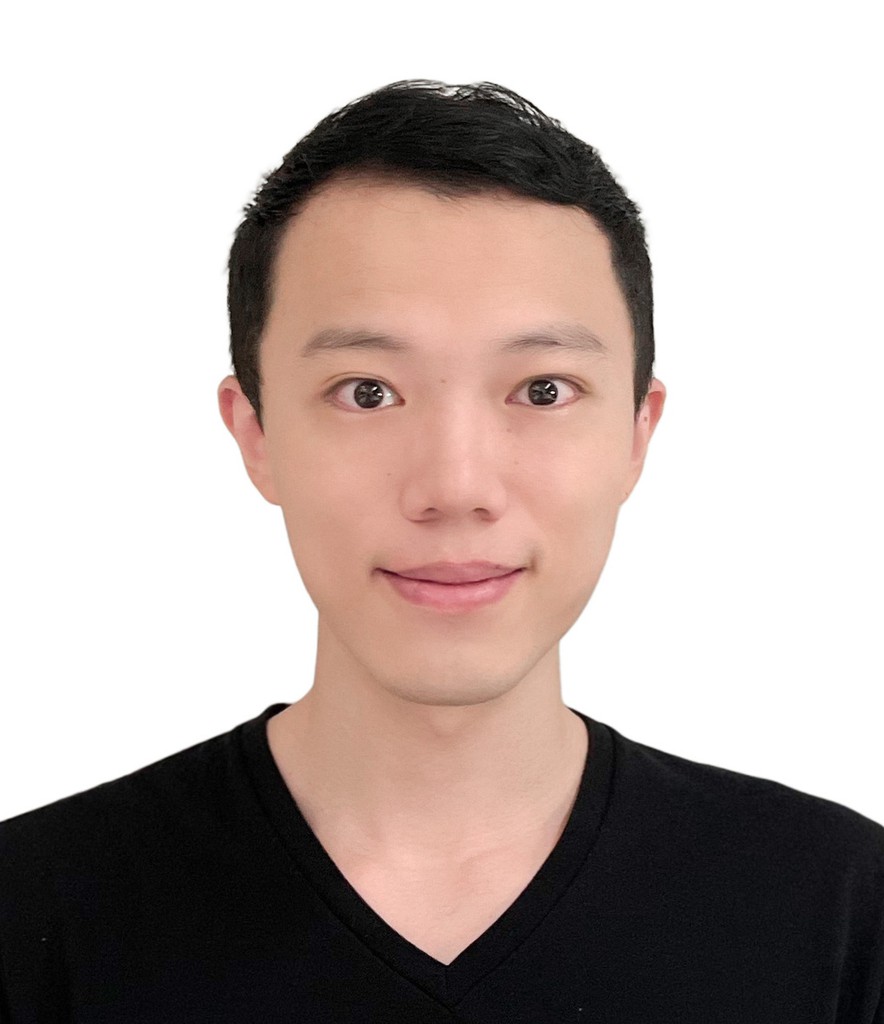}}]{Keqiang Sun}
Keqiang Sun is currently a Ph.D. candidate in electronic engineering from Chinese University of Hong Kong, advised by Hongsheng Li. His research focuses on 3D-aware generative model. He received his bachelor’s degree in electrical engineering from the Tianjin University and master's degree in integrated circuit engineering from Tsinghua University.
\end{IEEEbiography}
\vspace{-0.7cm}
\begin{IEEEbiography}[{\includegraphics[width=1in,height=1.25in,clip,keepaspectratio]{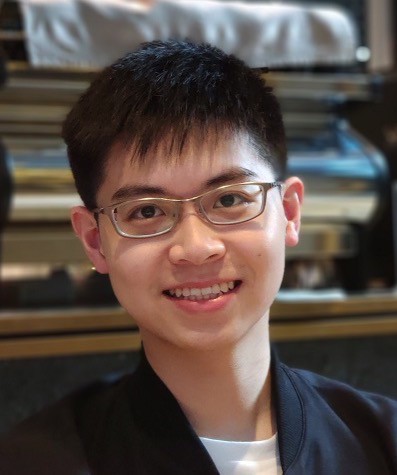}}]{Shangzhe Wu}
is a Postdoctoral Researcher in Stanford University.
He received his PhD from University of Oxford, supervised by Andrea Vedaldi and Christian Rupprecht, and bachelor’s degree from HKUST.
His current research focuses on unsupervised 3D perception and inverse rendering.
His work received the Best Paper Award at CVPR 2020.
\end{IEEEbiography}
\vspace{-0.5cm}
\begin{IEEEbiography}[{\includegraphics[width=1in,height=1.25in,clip,keepaspectratio]{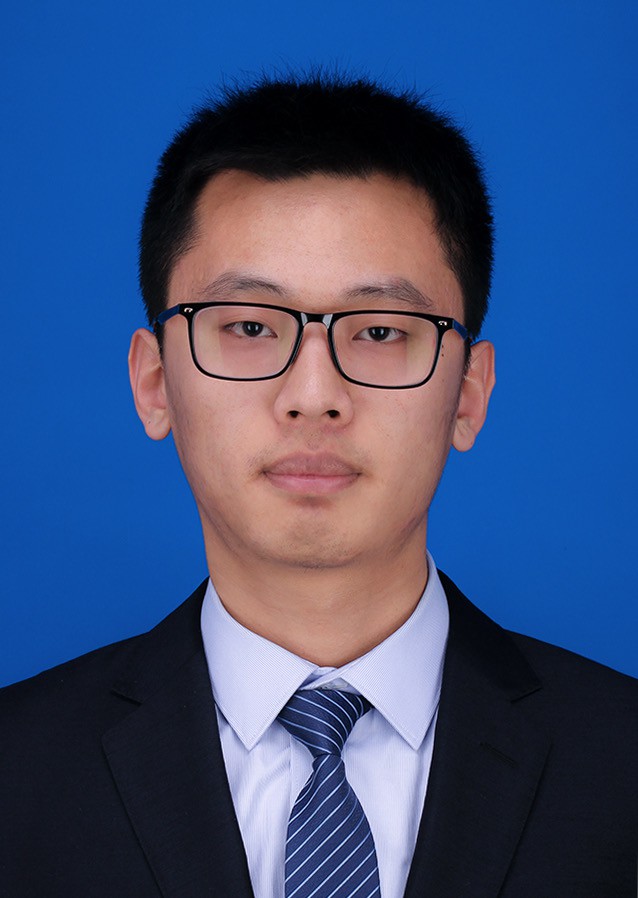}}]{Ning Zhang}
Ning Zhang received the B.S. and M.S. degrees in information engineering from Shanghai Jiao Tong University and integrated circuit engineering from Tsinghua University respectively. He is currently an algorithm engineer at Sensetime.  His research interests include computer vision and image synthesis.
\end{IEEEbiography}
\vspace{-0.5cm}
\begin{IEEEbiography}[{\includegraphics[width=1in,height=1.25in,clip,keepaspectratio]{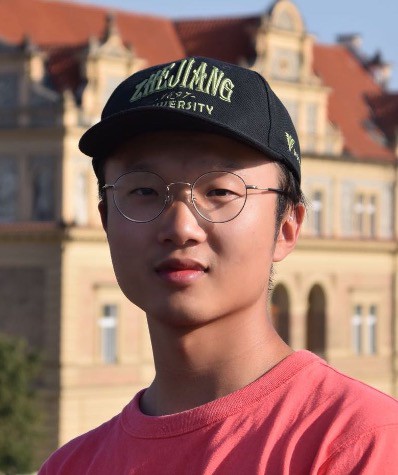}}]{Zhaoyang Huang}
Zhaoyang Huang received the bachelor and master's degrees in computer science from the Zhejiang University, where he worked with Guofeng Zhang and Xiaowei Zhou on visual localization and intrinsic image decomposition. He is currently working toward the DPhil degree with Multimedia Lab, CUHK, supervised by Hongsheng Li. His research focuses on correspondence learning and video editing.
\end{IEEEbiography}
\vspace{-0.5cm}
\begin{IEEEbiography}[{\includegraphics[width=1in,height=1.25in,clip,keepaspectratio]{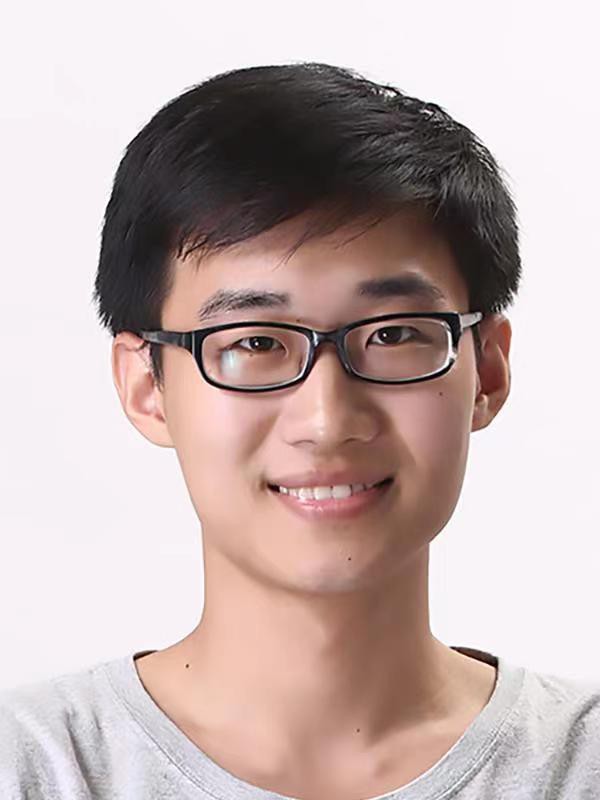}}]{Quan Wang}
Quan Wang received the B.S. degree in electronic engineering from Tsinghua University, China. He is currently a research scientist at Sensetime. His research interests include computer vision and 3D reconstruction.
\end{IEEEbiography}
\vspace{-0.5cm}
\begin{IEEEbiography}[{\includegraphics[width=1in,height=1.25in,clip,keepaspectratio]{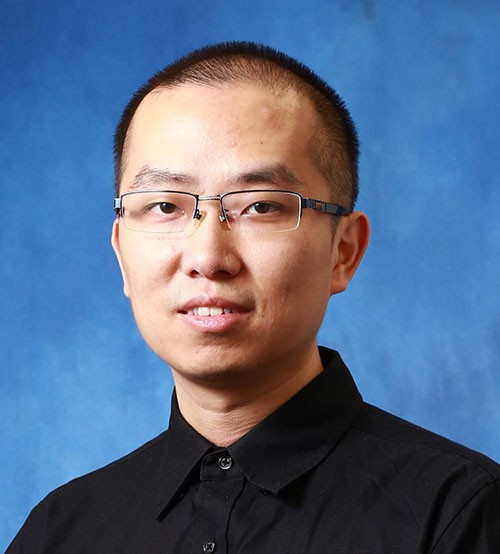}}]{Hongsheng Li}
Hongsheng Li (Member, IEEE) received the BS degree in automation from the East China University of Science and Technology, Shanghai, China, in 2006, and the MS and PhD degrees in computer science from Lehigh University, Bethlehem, Pennsylvania, in 2010 and 2012, respectively. He is currently an associate professor at the Department of Electronic Engineering, Chinese University of Hong Kong. His research interests include computer vision, medical image analysis, and machine learning.
\end{IEEEbiography}

\end{document}